\documentclass{article}

 \usepackage[preprint]{neurips_2025}

\usepackage[utf8]{inputenc} 
\usepackage[T1]{fontenc}    
\usepackage{hyperref}       
\usepackage{url}            
\usepackage{booktabs}       
\usepackage{amsfonts}       
\usepackage{nicefrac}       
\usepackage{microtype}      
\usepackage{xcolor}         
\usepackage[utf8]{inputenc} 
\usepackage[T1]{fontenc}    
\usepackage{url}            
\usepackage{booktabs}       
\usepackage{amsfonts}       
\usepackage{nicefrac}       
\usepackage{microtype}      
\usepackage{xcolor}         

\usepackage{color, colortbl}
\usepackage{soul} 
\usepackage{subcaption} 
\usepackage{caption}
\usepackage{graphicx, amsmath, multirow, wrapfig, float}

\title{EAGLE: Expert-Augmented Attention Guidance \\for Tuning-Free Industrial Anomaly Detection  \\in Multimodal Large Language Models}

\author{
Xiaomeng Peng \quad Xilang Huang \quad Seon Han Choi \vspace{2mm}\\
{Ewha Womans University}\\
 {\tt\small \{xiaomeng, seonhan.choi\}@ewha.ac.kr, \{huangxilang901\}@gcc.edu.cn}
}

\begin{document}

\maketitle

\begin{abstract}
\label{sec:abstract}
Multimodal large language models (MLLMs) can enrich industrial anomaly detection with semantic descriptions and anomaly reasoning, but they still lag specialist anomaly detectors in binary detection accuracy. Existing approaches address this gap by fine-tuning MLLMs or training bridging modules to align expert outputs with MLLM inputs, limiting flexibility across backbones. We propose EAGLE, a tuning-free framework that integrates expert anomaly detectors with frozen MLLMs. EAGLE consists of Threshold-Guided Prompt Selection (TGPS), which estimates a decision threshold from expert model statistics and selects textual and visual prompts, and Confidence-Aware Attention Sharpening (CAAS), which shifts MLLM attention toward visual evidence when expert confidence is low. Beyond improving accuracy, we analyze MLLM attention and find that correct anomaly predictions are associated with stronger focus on ground-truth defect regions; EAGLE consistently strengthens this alignment. On MVTec-AD and VisA, EAGLE improves five MLLM backbones without parameter updates, reaching up to 94.4\% and 88.1\% in anomaly discrimination accuracy, respectively, and achieving performance competitive with fine-tuning-based methods while largely preserving MLLM semantic reasoning ability. Code is available at \href{https://github.com/shengtun/Eagle}{https://github.com/shengtun/Eagle}.
\end{abstract}

\section{Introduction}
\label{sec:intro}
In intelligent manufacturing scenarios, Industrial Anomaly Detection (IAD) is a core task for ensuring product quality and operational safety. Although deep learning-based IAD models have achieved high performance on multiple benchmark datasets, they still suffer from a fundamental limitation in real-world deployment: their outputs are typically restricted to binary decisions, lacking interpretable semantic information such as anomaly type identification, precise localization, and descriptive explanations~\cite{jiang2024mmad}. This limitation makes it difficult to effectively support on-site troubleshooting and quality control. The emergence of MLLMs presents a new opportunity in this direction, owing to their strong visual understanding and language generation capabilities.

General MLLMs lack domain-specific knowledge of industrial defects, resulting in limited capability for anomaly detection tasks. To address this, existing works enhance MLLMs through various strategies: Some works construct high-quality instruction-tuning datasets for IAD, covering multi-level reasoning tasks ranging from visual defect description to root cause analysis, and use them to perform supervised fine-tuning (SFT) of MLLMs~\cite{gu2024anomalygpt, li2023myriad, xu2025towards}; others introduce chain-of-thought reasoning frameworks with structured output formats to enforce step-by-step analytical reasoning before reaching a final decision~\cite{zeng2025lr, chao2025anomalyr1, zhao2025omniad}; more recent works further incorporate Group Relative Policy Optimization (GRPO) with customized reward functions, such as difficulty-aware reweighting and reasoning-outcome alignment metrics, to improve the quality of model reasoning~\cite{guan2025emit, chao2025anomalyr1, zhao2025omniad}.

On the other hand, the visual encoders of general MLLMs are pretrained with a primary focus on global semantic alignment, making them less responsive to localized, fine-grained visual irregularities~\cite{tong2024eyes}. To overcome this, existing methods typically introduce extra expert models to inject visual cues and textual prompts into MLLMs. Since the anomaly maps produced by expert models are too abstract for MLLMs to interpret directly, these methods commonly rely on fine-tuning or additional trainable modules to bridge the gap between the two. Although these methods have achieved certain progress in anomaly localization and anomaly understanding, fine-tuning-based frameworks lack flexibility, as replacing the MLLM backbone requires retraining the entire system. More critically, despite the substantial resources devoted to fine-tuning, the binary anomaly classification performance of these methods still falls considerably short of traditional deep learning models.

To overcome these limitations, we propose EAGLE, a tuning-free framework that synergizes expert models with MLLMs to achieve high detection accuracy while enabling semantic anomaly analysis. Unlike prior works~\cite{gu2024anomalygpt, li2023myriad, li2025triad, xu2025towards} that adapt MLLMs to expert model outputs through additional training modules or fine-tuning, EAGLE introduces \textbf{Threshold-Guided Prompt Selection (TGPS)}, which selectively converts anomaly scores produced by expert models into textual and visual prompts suitable for MLLMs, thereby effectively guiding MLLMs toward accurate anomaly detection.

We further note that MLLMs exhibit a well-known limitation in their attention allocation, where the ratio of attention to textual tokens over visual tokens can reach as high as 9:1, causing MLLMs to prioritize linguistic information over visual evidence~\cite{woo2025don, chen2025spatial, An_2025_CVPR}. We observe that when expert models produce erroneous predictions, the resulting misleading textual prompt can override visual evidence in later transformer layers, even when MLLMs already attend to ground-truth anomalous regions in intermediate layers, ultimately leading to incorrect predictions. To address this, we introduce \textbf{Confidence-Aware Attention Sharpening (CAAS)}, which selectively amplifies visual attention when expert predictions are uncertain, enabling MLLMs to rely more on visual evidence and mitigating hallucinations caused by unreliable textual prompt.

\begin{wrapfigure}{r}{0.5\textwidth} 
  \centering
   \vspace{-10pt}
  \includegraphics[width=\linewidth]{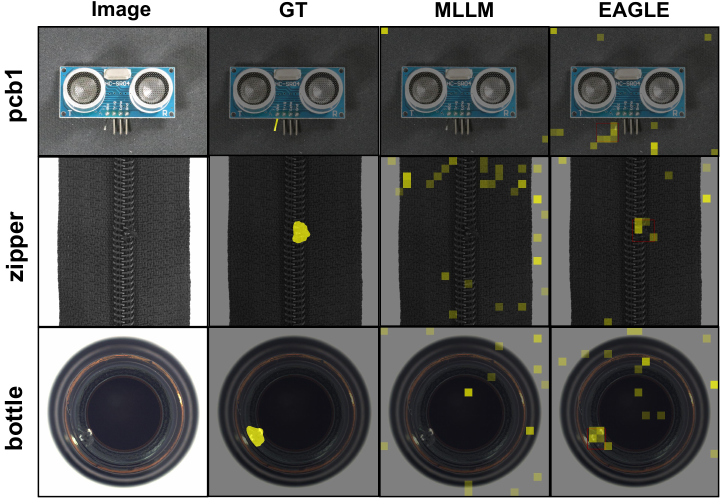}
  \vspace{-10pt}
  \caption{Attention map visualizations from Qwen2.5-VL-7B on the MVTec-AD and VisA. }
  \label{fig:2}
\vspace{-15pt}
\end{wrapfigure}

Our experimental analysis further shows that samples with correct predictions tend to exhibit higher attention concentration on ground-truth defect regions, indicating a strong correlation between attention alignment and prediction accuracy. As illustrated in Fig.~\ref{fig:2}, EAGLE is able to redirect the attention of MLLMs toward anomalous regions, suggesting improved focus on relevant visual evidence during answer generation. 

Experimental results on industrial anomaly detection benchmarks, including MVTec-AD~\cite{bergmann2019mvtec}, VisA~\cite{zou2022spot}, RAD~\cite{cheng2024rad}, BTAD~\cite{mishra21-vt-adl}, and MPDD~\cite{jezek2021deep}, demonstrate that EAGLE consistently improves the performance of multiple MLLMs and achieves competitive or superior results to fine-tuning-based methods in terms of anomaly discrimination accuracy and F1 score.
Furthermore, experiments on the multimodal benchmark MMAD~\cite{jiang2024mmad}, which incorporates both visual and textual information, further demonstrate the generalizability of EAGLE across diverse anomaly detection tasks.


\section{Related work}
\label{sec:formatting}
\subsection{Industrial Anomaly Detection}
Traditional deep learning-based IAD methods fall into two main categories. \textbf{Reconstruction-based methods}~\cite{Recon_1, jiang2024toward} train generative models to reconstruct normal images. During testing, anomalous samples differ from normal patterns and therefore cannot be accurately reconstructed, leading to higher reconstruction errors that are used to identify anomalies. \textbf{Feature embedding methods} leverage pretrained visual encoders to extract image features, employing memory banks~\cite{roth2022towards, Feture_embed_3} or student-teacher architectures~\cite{Feture_embed_efficientad, Feture_embed_ST} to model the feature space of normality. Anomalies are identified by measuring the deviation of test samples from normal data in the feature space.
Recently, vision-language models have shown promise for IAD. Methods such as WinCLIP~\cite{jeong2023winclip}, AnomalyCLIP~\cite{zhou2023anomalyclip}, and PromptAD~\cite{li2024promptad} exploit pretrained models like CLIP for few-shot and zero-shot anomaly detection. However, all these methods provide limited explanations and interpretability for detected anomalies.

\subsection{ Multimodal Large Language Model}
The emergence of MLLMs has introduced new opportunities for IAD, extending the capability of detection systems beyond binary decisions to encompass richer outputs such as semantic descriptions and in-depth anomaly analysis. Despite this promise, general MLLMs exhibit two fundamental limitations when applied to IAD tasks. However, general MLLMs are pretrained on broad vision-language objectives, such as image captioning and visual question answering, which do not expose the model to the specialized characteristics of industrial defects. As a result, these models demonstrate limited capability in accurately understanding and describing anomalies. Considerable research effort has been devoted to addressing this gap. AnomalyGPT~\cite{gu2024anomalygpt}, Myriad~\cite{li2023myriad}, and Anomaly-OV~\cite{xu2025towards} construct high-quality instruction-tuning datasets to inject domain-specific anomaly knowledge into MLLMs. LR-IAD~\cite{zeng2025lr} and AnomalyR1~\cite{chao2025anomalyr1} introduce chain-of-thought reasoning and structured inference frameworks to enhance step-by-step analytical capability. EMIT~\cite{guan2025emit} and OmniAD~\cite{zhao2025omniad} further incorporate GRPO with customized reward functions to improve the quality of model reasoning.

Furthermore, the visual encoders of general MLLMs are pretrained with a primary focus on global semantic alignment, such as image-text matching and scene-level understanding, making them less responsive to localized, fine-grained visual irregularities~\cite{tong2024eyes} that characterize industrial defects. To overcome this, several works integrate dedicated visual encoders or expert models to strengthen anomaly perception. AnomalyGPT~\cite{gu2024anomalygpt} introduces a dedicated image decoder coupled with a prompt learner to align pixel-level anomaly localization with the LLM. Myriad~\cite{li2023myriad} designs a visual-expert-guided vision encoder and converts expert outputs into LLM-compatible textual prompts via a dedicated prompt generator. Anomaly-OV~\cite{xu2025towards} proposes a Look-Twice Feature Matching mechanism equipped with a learnable visual token selection module to identify the most discriminative anomaly features.

Despite these advances, existing methods generally require substantial annotated data and considerable computational resources for fine-tuning. More critically, on the fundamental task of binary anomaly classification, their performance still falls considerably short of traditional deep learning models. In contrast, we propose EAGLE, a tuning-free framework that bridges this performance gap by elevating binary classification accuracy to a level comparable to that of deep learning models, while simultaneously preserving the semantic analysis capabilities inherent to MLLMs.

\section{Method}

\subsection{Preliminary}

\subsubsection{Memory-based anomaly detection}
The expert model adopts the PatchCore~\cite{roth2022towards} architecture. Following its standard procedure, a pre-trained feature extractor (e.g., WideResNet50) is used to extract feature maps $\mathcal{F}_i = \phi_i \in \mathbb{R}^{C \times H \times W}$ from each training image $x_i$. Patch features from all training samples are aggregated into a patch feature set $\mathcal{F}$, from which a memory bank $\mathcal{M}$ is constructed via a coreset sampling algorithm $\mathcal{G}$.
\begin{equation}
\label{eq:memory_bank}
\mathcal{M} = \mathcal{G}(\mathcal{F}), \quad \mathcal{M} \subset \mathcal{F}
\end{equation}

In the inference stage, given a test image $x_t$, patch-level features $\mathcal{F}_t$ are extracted using the same feature extractor as in training. For each patch feature $f_t^{(h,w)} \in \mathcal{F}_t$, its anomaly score $s_t^{(h,w)}$ is defined as the Euclidean distance to its nearest neighbor in the memory bank $\mathcal{M}$:
\begin{equation}
\label{max_distance}
m^* = \arg\min_{m \in \mathcal{M}} \| f_t^{(h, w)} - m \|_2 ,
\end{equation}
\begin{equation}
\label{max_distance_2}
s_t^{(h, w)} = \| f_t^{(h, w)} - m^* \|_2 .
\end{equation}
Here, $(f_t^{(h, w)}, m^*)$ denotes a test patch and its corresponding nearest neighbor in $\mathcal{M}$. The image-level anomaly score $s_{img}$ is defined as the maximum patch-level score over all spatial locations:
\begin{equation}
s^t_{img} = \max(s_t^{(1,\: 1)} ,s_t^{(1,\: 2)}, \ldots, s_t^{(h, \: w)}).
\label{eq:image_score}
\end{equation}

\subsubsection{Extreme Value Theory}
\label{EVT}
The goal of extreme value theory (EVT) is to model the statistical behavior of extreme events. 
A classical result established by Fisher and Tippett~\cite{fisher1928limiting} and later formalized by Gnedenko~\cite{gnedenko1943distribution} 
states that if $X_1, \dots, X_n$ are independent and identically distributed (i.i.d.) random variables with a common distribution function $F$, 
and
\begin{equation}
X_{(n)} = \max(X_1, \ldots, X_n),
\label{eq:evt}
\end{equation}

denotes their maximum, then under mild regularity conditions, the properly normalized maximum converges in distribution to a limiting distribution belonging to the Generalized Extreme Value (GEV) family.
The cumulative distribution function of the GEV distribution is given by
\begin{equation}
G_{\gamma}(x) =
\exp\left[-(1+\gamma x)^{-1/\gamma}\right],
\quad \gamma \in \mathbb{R}, \quad 1+\gamma x > 0,
\end{equation}

where $\gamma$ is the extreme value index controlling the tail behavior. All extreme values of common standard distributions follow one of these forms of $G_{\gamma}(x)$, depending on their original distribution $F(x)$.

\begin{figure*}[t]
    \centering
    \includegraphics[width=\linewidth]{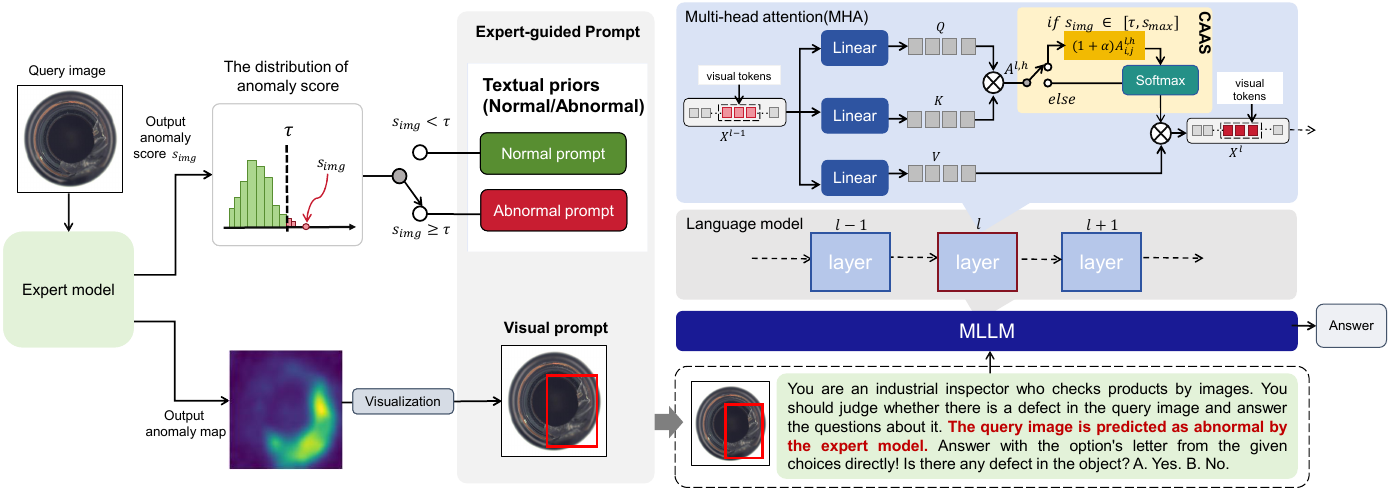}
    \caption{
    \textbf{Pipeline of EAGLE.} Given a query image, an expert model produces an image-level anomaly score $s_{img}$ and an anomaly map. The anomaly score is compared against a threshold $\tau$ to select the appropriate prompt. The MLLM incorporates the proposed CAAS module to amplify attention on visual tokens, ultimately generating the final answer.
    }
    \label{fig:3}
\vspace{-8pt}
\end{figure*}

\subsection{Framework Overview}
As illustrated in Figure~\ref{fig:3}, we propose EAGLE, a tuning-free framework for anomaly detection. To avoid fine-tuning overhead, we introduce two key mechanisms that transform the anomaly scores produced by the expert model into prompt forms that are easily interpretable by MLLMs. TGPS (Sec.~\ref{TGPS}) determines a decision threshold by estimating the anomaly score distribution of normal samples, which is then used to classify test samples and determine the corresponding visual and textual prompts as input to MLLMs. The CAAS mechanism (Sec.~\ref{CAAS}) identifies low-confidence predictions from the expert model and selectively amplifies visual token attention in the intermediate layers of MLLMs. By strengthening visual evidence, CAAS prevents the MLLMs from over-relying on erroneous linguistic priors, enabling more accurate correction of expert model misclassifications.
\begin{figure}[t]
    \centering
    \includegraphics[width=\linewidth]{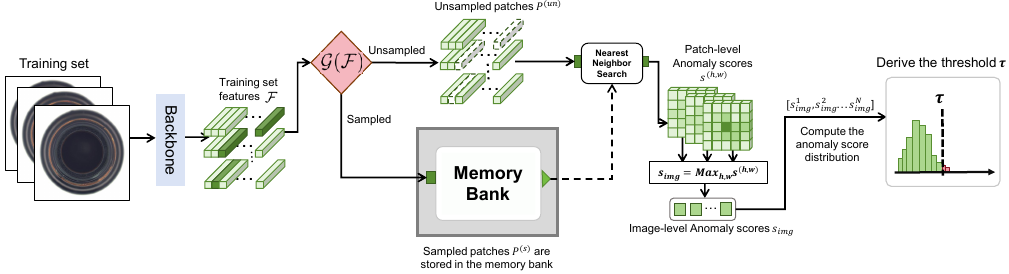}
    \vspace{-10pt}
    \caption{
    Normal training images are converted into patch-level features $\mathcal{F}$, which are stored in a memory bank constructed via greedy coreset sampling $\mathcal{G}$. For each training image $x_i$, its unsampled patch set $P_i^{(un)}$ is used to compute the image-level anomaly score through nearest-neighbor search. Finally, the anomaly score distribution of the training set is used to estimate the threshold $\tau$.
    }
    \label{fig:4}
    \vspace{-10pt}
\end{figure}

\subsection{Threshold-Guided Prompt Selection}
\label{TGPS}

\subsubsection{Distribution-Based Threshold selection}

In PatchCore, anomaly scores are computed as distances between test patch features and the memory bank, while the threshold is typically selected manually. To automatically compute thresholds before testing, we propose the TGPS mechanism, which estimates image-level decision thresholds by modeling the distribution of anomaly scores using unsampled patch features during memory bank construction. Fig.~\ref{fig:4} provides an overview of the TGPS pipeline.



\begin{wrapfigure}{r}{0.5\textwidth}
\centering
\begin{minipage}{0.49\linewidth}
\centering
\includegraphics[width=\linewidth]{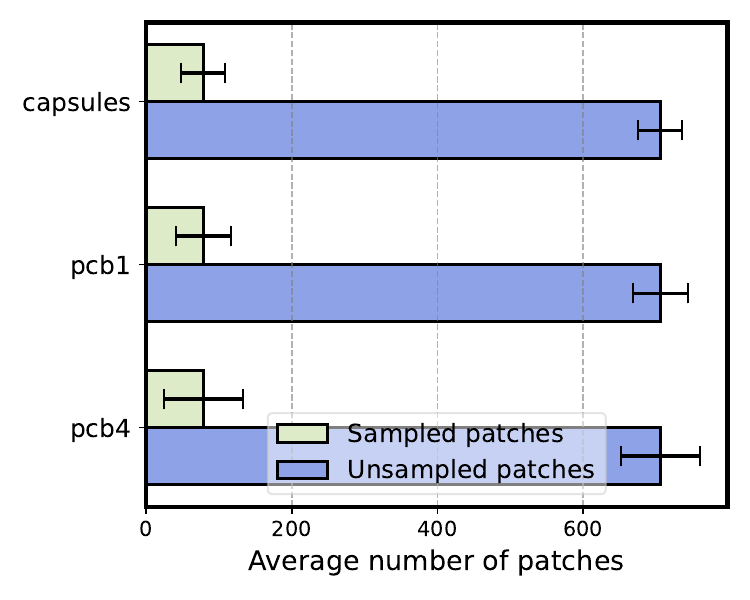}
\captionof{figure}{Sampled and unsampled patches.}
\label{fig:5}
\end{minipage}
\hfill
\begin{minipage}{0.49\linewidth}
\centering
\includegraphics[width=\linewidth]{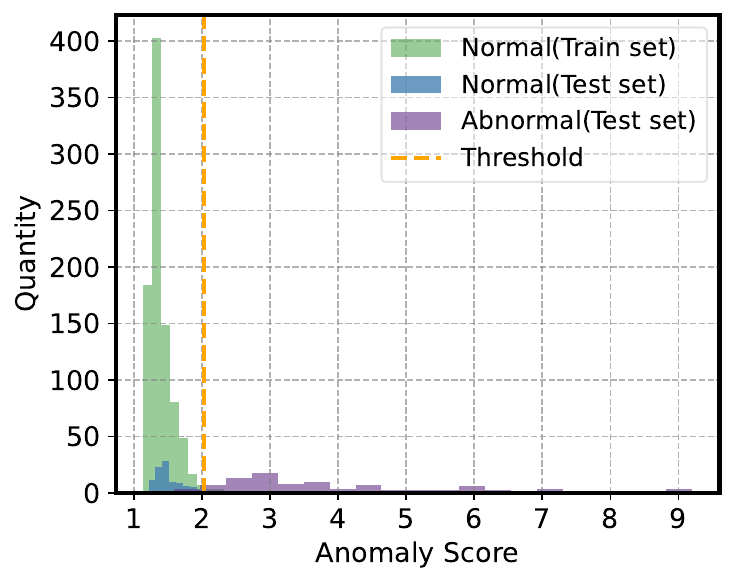}
\captionof{figure}{AS distribution of PCB1.}
\label{fig:6}
\end{minipage}
\vspace{-15pt}
\end{wrapfigure}
A key observation motivating TGPS is that only a small fraction of patch features are retained in the memory bank, while the majority are discarded during coreset sampling. As illustrated in Fig.~\ref{fig:5}, we analyze the proportion of sampled and unsampled patches per training image during memory bank construction on representative classes. The results show that, on average, only about 10\% of patches per image are selected into the memory bank, while the unsampled patches account for approximately 90\% of all extracted features per image.

Formally, for each training image $x_i \in X^{\text{train}} = \{x_1, x_2, \ldots, x_N\}$, where all samples are normal, we denote the set of sampled patch features as $P_i^{(s)}$ and the set of unsampled patch features as
\begin{equation}
\label{max_distance_3}
\
P_i^{(un)} = \mathcal{F}_i \setminus P_i^{(s)},
\end{equation}
where $\mathcal{F}_i = \{ f_i^{(h, w)} \}$ represents the complete set of patch features extracted from $x_i$.

Following the anomaly score formulation (Eqs.~\ref{max_distance}, \ref{max_distance_2}, 
and \ref{eq:image_score}), the image-level anomaly score is computed as the maximum over patch-level scores. Intuitively, the patch $f_i^*$ that attains the maximum anomaly score is unlikely to be included in $P_i^{(s)}$, since if $f_i^*$ were sampled into $\mathcal{M}$, its anomaly score would reduce to the distance between itself and an identical entry, i.e., $\|f_i^* - m^*\|_2 = 0$ with $m^* = f_i^*$. Therefore, for training sets, the image-level anomaly score typically arises from $P_i^{(un)}$ .

As a result, we can compute the image-level anomaly scores for all normal training samples and denote them as $\mathcal{S} = \{ s^1_{img}, s^2_{img}, \ldots, s^N_{img} \}$. By the Fisher-Tippett-Gnedenko theorem (Sec.~\ref{EVT}), the image-level 
anomaly score, computed as the maximum over all patch-level scores, naturally 
corresponds to the extreme value statistic (Eq.~\ref{eq:evt}), suggesting 
that the distribution of $\mathcal{S}$ converges to an extreme value 
distribution $G_{\gamma}(x)$. 
To estimate the decision threshold $\tau$, a GEV distribution is fitted to the image-level anomaly score set $\mathcal{S}$, and the threshold is defined as:
\begin{equation}
    \tau = F_{\text{GEV}}^{-1}(q),
\end{equation}
where $F_{\text{GEV}}^{-1}$ denotes the inverse cumulative distribution function of the fitted GEV distribution and $q$ is the confidence factor.

As shown in Fig.~\ref{fig:6}, the green histogram corresponds to the anomaly scores of normal training images, while the blue histogram represents those of normal test images. We observe that the distribution estimated from the training set covers most normal test samples, owing to the large number of normal training images. In contrast, images containing defects produce substantially higher anomaly scores, as defect regions yield large patch-level distances that exceed the normal score distribution, as illustrated by the purple histogram in Fig.~\ref{fig:6}. 
More discussions on TGPS applicability and limitations are provided in Appendix~\ref{app:AS}, while more anomaly score distributions are shown in Fig.~\ref{app:as-mvtec}.


\subsubsection{Conditional Prompt Selection for MLLMs}

During inference, the expert model provides two types of guidance to the MLLMs in the form of visual and textual prompts. More detailed descriptions are provided in Appendix~\ref{app:prompt}.

\textbf{Visual prompt}: We upsample the patch-level anomaly scores obtained from Eqs.~\ref{max_distance},~\ref{max_distance_2} to the original image resolution to construct an anomaly map. Based on this map, potential defect regions are localized and highlighted by drawing red bounding boxes on the input image. Since the expert model may also highlight regions in normal images, these visual cues can mislead the MLLMs. Therefore, visual prompts are provided only for images predicted as anomalous. \textbf{Textual prompt}: For each test image, we compare its image-level anomaly score $s^t_{img}$ (Eq.~\ref{eq:image_score}) with the threshold $\tau$ to select the corresponding textual prior. When $s^t_{img} < \tau$, the prior reads ``\textit{This image is predicted as normal.}''; otherwise, it is set to ``\textit{This image is predicted as abnormal.}'' . This design transforms the quantitative judgment of the expert model into a natural language signal that the MLLM can readily comprehend.

\subsection{MLLMs with Confidence-Aware Attention Sharpening}
\label{CAAS}
Since PatchCore relies on local patch-level feature comparisons, it tends to produce false positives on certain datasets, misclassifying normal textural patterns as anomalies (See Sec.~\ref{sec:attn} for more details). Moreover, prior work~\cite{chen2025spatial, yin2025clearsight, leng2024mitigating, wang2024mllm, kang2025see} has shown that MLLMs allocate significantly more attention to textual tokens than visual tokens, making them susceptible to erroneous textual priors (see Appendix~\ref{app:2}). Consequently, when the expert model mispredicts, the MLLM may be misled even when visual anomaly cues are present. Accordingly, we introduce CAAS, which enhances MLLM attention weights to visual prompt tokens when the expert model is uncertain, thus reducing the influence of incorrect textual priors. This is supported by the preliminaries in Appendix~\ref{app:1}.

We define the low-confidence region as $s_{img}^t \in [\tau, s_{\text{max}}]$, where $\tau$ is the threshold estimated by TGPS and $s_{\text{max}}$ is the maximum anomaly score of normal training samples. This interval corresponds to the overlap between normal and abnormal distributions, where the expert model has high uncertainty. CAAS is activated only within this region. If $\tau > s_{\text{max}}$, CAAS is disabled.

Based on prior works \cite{chen2025spatial, yin2025clearsight} showing that intermediate layers are highly sensitive to visual reasoning, we selectively adjust attention allocation only in intermediate layers. Additional analysis on intermediate layers is provided in the Appendix~\ref{appendix_caas}. To influence the attention weights while preserving the normalized attention distribution, we adjust the attention logits before the softmax operation. For the attention logit $A_{i,j}^{(l,h)}$ in layer $l$ and head $h$, the adjusted logit is computed as:
\begin{equation}
A_{i,j}^{(l,h)} = \begin{cases}
(1+\alpha) \cdot \mathbf{A}_{i,j}^{(l,h)} & \text{if } j \in \mathcal{I} \\
\mathbf{A}_{i,j}^{(l,h)} & \text{otherwise}
\end{cases}, \mathrm{if }\,\, s_{img}^t \in [\tau, s_{\text{max}}],  
\end{equation}
where $\mathcal{I}$ represents the index set of visual prompt tokens. $\alpha $ is the scaling factor.

\section{Experiments}
\subsection{Experiment Details}
\label{sec:experiment-1}

\noindent\textbf{Datasets.} We evaluate our framework on multiple benchmarks: 
MVTec-AD~\cite{bergmann2019mvtec}, VisA~\cite{zou2022spot}, RAD~\cite{cheng2024rad} and MMAD benchmark~\cite{jiang2024mmad}.  with additional datasets results in Appendix~\ref{app:addition_results}.

\noindent\textbf{Evaluation Metrics.} We report Precision, Recall, F1-score, 
and Accuracy$^*$. Detailed definitions are provided in Appendix~\ref{app:metrics}. 

\noindent\textbf{Implementation Detail.} We adopt PatchCore~\cite{roth2022towards} with WideResNet-50 as the backbone. Image preprocessing details are provided in Appendix~\ref{app:implementation detail}. In TGPS, we use $q=0.98$. For CAAS, we set $\alpha=0.6$ and apply attention modulation to intermediate layers ($9 \leq l \leq 14$).

\textbf{Baseline.} We compare our approach with existing fine-tuning-based and GRPO-based methods. The fine-tuning-based methods include specialized MLLMs for anomaly detection \cite{gu2024anomalygpt, li2023myriad}, as well as MLLMs trained via GRPO and fine-tuning on anomaly detection datasets\cite{zeng2025lr, zhao2025omniad}.

\textbf{MLLM backbones.} We evaluate EAGLE on five diverse MLLMs, including LLaVA-1.5-7B\cite{liu2024improved}, LLaVA-NeXT-7B\cite{liu2024llavanext}, Qwen2.5-VL-7B\cite{bai2025qwen2}, InternVL3-8B\cite{zhu2025internvl3}, MiniCPM-8B\cite{hu2024minicpm}. 

\begin{table*}[!htbp]
\centering
\vspace{-5pt}
\caption{Quantitative comparison between MLLMs and MLLMs with EAGLE on the MVTec-AD, VisA, and RAD datasets. Baseline MLLMs follow a 1-shot$^+$ template image setting(see Appendix~\ref{app:implementation detail}). Accuracy$^*$ denotes Balanced Accuracy, computed as the average of true positive rate and true negative rate, consistent with the Anomaly Discrimination metric in MMAD~\cite{jiang2024mmad}. }
\vspace{-5pt}
\label{tab:mvtec_visa}
\resizebox{\linewidth}{!}
{%
\begin{tabular}{l c cccc cccc cccc}
\toprule[0.4mm]
\multirow{2}{*}{\textbf{Method}} &\multirow{2}{*}{ \textbf{Scale} }
& \multicolumn{4}{c}{\textbf{MVTec-AD}} 
& \multicolumn{4}{c}{\textbf{VisA}} 
& \multicolumn{4}{c}{\textbf{RAD}} 
\\
\cmidrule(lr){3-6}\cmidrule(lr){7-10}\cmidrule(lr){11-14}
& 
& \textbf{Accuracy$^*$} & \textbf{Precision} & \textbf{Recall} & \textbf{F1}
& \textbf{Accuracy$^*$} & \textbf{Precision} & \textbf{Recall} & \textbf{F1}
& \textbf{Accuracy$^*$} & \textbf{Precision} & \textbf{Recall} & \textbf{F1} \\
\midrule
PatchCore          & - &92.1 & 96.7 & 96.3 & 96.4 
                & 88.0 & 96.1 & 79.7   & 85.4  
                & 94.6 & 98.6 & 94.8 & 96.5
\\
\midrule
LLaVA-1.5          & 7B & 61.6 & 79.2 & 72.3 & 71.9 & 58.4  &  75.0 & 26.5 &  32.8  
& 62.1 & 88.4 & 31.0 & 40.3
\\
\rowcolor{gray!20}
\textbf{LLaVA-1.5  w/ours}   & 7B & \textbf{93.0} & \textbf{95.6} & \textbf{96.6} & \textbf{96.0}
&\colorbox{yellow!20}{\textbf{88.1}} & \textbf{96.1} & \textbf{79.8} & \textbf{85.4}
& \textbf{94.6} & \textbf{98.6} & \textbf{94.8} & \textbf{96.5}
\\
\midrule
LLaVA-NeXT        & 7B & 69.8 & \textbf{99.5} & 40.2 & 50.9 & 59.0 & 80.8 & 18.8 & 27.9
&83.7 &\textbf{100.0} & 67.4 & 78.3
\\
\rowcolor{gray!20}
\textbf{LLaVA-NeXT w/ours}   & 7B & \textbf{92.1} & 96.6 & \textbf{95.7} &\textbf{96.1} 
&\colorbox{yellow!20}{\textbf{88.1}} &\textbf{95.2}  & \textbf{80.5} &\textbf{88.6} 
&\textbf{89.2} &\textbf{95.6}  & \textbf{97.2} &\textbf{96.3} 
\\
\midrule
MiniCPM-V4.5           & 8B & 62.7&79.8  & 38.0&45.6  &62.7  & 79.8 & 38.0 & 45.6 &72.1& 88.0 & 52.5 & 60.1\\
\rowcolor{gray!20}
\textbf{MiniCPM-V4.5  w/ours}   & 8B & \textbf{92.9} & \textbf{95.7} & \textbf{96.6} & \textbf{96.0}
& \textbf{86.5} & \textbf{95.4} & \textbf{77.4} & \textbf{83.6}
& \textbf{88.4} & \textbf{97.0} & \textbf{88.2} & \textbf{92.2}
\\
\midrule
InternVL3          & 8B & 83.9 & 91.2 & 89.7 & 90.0 & 74.2 &89.6  &58.6  &66.1
&78.1 &90.8 &97.2 &93.9
\\
\rowcolor{gray!20}
\textbf{InternVL3 w/ours}   & 8B & \textbf{93.0} & \textbf{95.7} & \textbf{96.6} & \textbf{96.0}
& \textbf{87.9} & \textbf{96.1} & \textbf{79.5} & \textbf{85.3}
& \colorbox{yellow!20}{\textbf{95.0}} & \textbf{99.0} & \textbf{94.2} & \textbf{96.4}
\\
\midrule
Qwen2.5-VL           &7B &85.9 &95.0 & 83.9 & 88.3&76.0&  89.2& 61.2 & 69.4 
&82.8 &94.8 &83.5 &87.8 \\
\rowcolor{gray!20}
\textbf{Qwen2.5-VL w/ours}   & 7B & \colorbox{yellow!20}{\textbf{94.4}} & \textbf{97.5} & \textbf{94.9} & \textbf{96.1} & 
\textbf{86.5} & \textbf{96.0} & \textbf{76.6} & \textbf{83.3}  
& \textbf{92.1} & \textbf{99.9} & \textbf{84.5} & \textbf{91.3}

\\
\bottomrule[0.4mm]
\end{tabular}
}
\end{table*}
\begin{table}[t]
\centering
\vspace{-10pt}
\caption{Comparison of performance on the VisA and MVTec datasets within the MMAD benchmark.}
\label{tab:mmad}
\resizebox{\textwidth}{!}{%
\begin{tabular}{lcccccccc}
\toprule[0.3mm]
\multirow{2}{*}{\textbf{Method}} 
& \multicolumn{1}{c}{\textbf{Anomaly}} 
& \multicolumn{4}{c}{\textbf{Defect}} 
& \multicolumn{2}{c}{\textbf{Object}} 
& \multirow{2}{*}{\textbf{Average} }\\

\cmidrule(lr){2-2} \cmidrule(lr){3-6} \cmidrule(lr){7-8}
& \textbf{Discrimination} 
& \textbf{Classification} 
& \textbf{Localization} 
& \textbf{Description} 
& \textbf{Analysis} 
& \textbf{Classification} 
& \textbf{Analysis} 
& \\
\midrule
InternVL3    & 79.1 & \textbf{59.9} & 63.8 & \textbf{75.7} & 83.0 & \textbf{82.6} & \textbf{88.4} & 76.1 \\
\rowcolor{gray!20}
\textbf{EAGLE}    & \textbf{90.5} & 59.5 & \textbf{66.6} & 73.5 & \textbf{83.5} & 80.3 & 87.0& \textbf{77.3}\\
\bottomrule[0.3mm]
\end{tabular}
}
\vspace{-5pt}
\end{table}

\subsection{Quantitative Results}

\textbf{Performance on industrial datasets}
Tab.~\ref{tab:mvtec_visa} reports the performance of EAGLE across different MLLM backbones on the MVTec-AD, VisA, and RAD datasets. Highlighted values indicate the best performance for each dataset. EAGLE leads to reliable performance gains for all evaluated backbones across all datasets. In particular, many baseline MLLMs exhibit relatively low recall, which can be attributed to their tendency to under-detect anomalies when visual evidence is weak or ambiguous. By introducing TGPS mechanism to control the injection of prompts, EAGLE improves recall in a stable manner, thereby enhancing overall anomaly detection performance. 

\begin{wraptable}{r}{0.5\textwidth}
\vspace{-10pt}
    \centering
    \caption{Comparison of our method with existing approaches across training modes. Performance is evaluated using standard classification accuracy.}
    \vspace{-2pt}
    \resizebox{1\linewidth}{!}
    {
    \begin{tabular}{clccc}
        \toprule[0.5mm]
        Mode & {{Method}} & MVTec-AD & VisA \\
        \midrule

       \multirow{2}[1]{*}{Fine-tuning} & AnomalyGPT       & 86.1 $\pm$ 1.1 & 77.4 $\pm$ 1.0   \\
        & Myriad      & 87.4 $\pm$ 0.9 & 80.0 $\pm$ 0.4   \\ 
        \midrule
         \multirow{2}[1]{*}{GRPO \& Fine-tuning} &LR-IAD        & 84.4 $\pm$ 0.0 & \textbf{87.6 $\pm$ 0.0}     \\
        &OmniAD  & \textbf{96.0 $\pm$ 0.0} & 86.6 $\pm$ 0.0  \\ 
        \midrule
        \multirow{2}[1]{*}{Tuning-free} 
        & \cellcolor{gray!20}\textbf{EAGLE (LLaVA-1.5)} & \cellcolor{gray!20}94.4 $\pm$ 0.0 & \cellcolor{gray!20}\underline{86.9 $\pm$ 0.0}  \\
       & \cellcolor{gray!20}\textbf{EAGLE (Qwen2.5-VL)} & \cellcolor{gray!20}\underline{94.7 $\pm$ 0.0} & \cellcolor{gray!20}85.1 $\pm$ 0.0  \\
        \bottomrule[0.5mm]
    \end{tabular}}
    \label{tab:comparison_mllm}
    \vspace{-14pt}
\end{wraptable}

Notably, EAGLE operates in a tuning-free manner without any task-specific parameter 
updates. As shown in Tab.~\ref{tab:comparison_mllm}, EAGLE achieves competitive performance against existing fine-tuning-based and GRPO-based approaches across both MVTec-AD and VisA, while maintaining a fully tuning-free setting. In particular, EAGLE demonstrates stable performance across different MLLM backbones without introducing additional trainable modules.



\textbf{Performance on MMAD.} Tab.~\ref{tab:mmad} reports the average performance on the MVTec-AD and VisA subsets of MMAD, while the full MMAD results are provided in the Tab.~\ref{tab:app mmad}. EAGLE outperforms the baseline MLLM in several anomaly-aware tasks, achieving notable gains in Anomaly Discrimination (+11.4\%), Defect Localization (+2.8\%), and Defect Analysis (+0.5\%). Slight performance degradation is observed in some other tasks that rely more on global image understanding.


\subsection{Ablation Studies}

We validate each module in the EAGLE on Qwen2.5-VL-7B\cite{bai2025qwen2} through ablation studies. 

\textbf{Impact of Expert model.}
Tab.~\ref{tab:prompt_design} compares performance under different prompting 
modalities. Using either visual or textual prompts alone consistently degrades 
performance, suggesting that neither modality alone is sufficient and that the 
expert model contributes beyond its final prediction. The complete TGPS, 
combining both modalities, achieves consistent improvements, validating their 
complementarity. Notably, these improvements do not merely reflect the MLLM 
echoing the expert's prediction; rather, both prompts actively guide the MLLM 
to focus attention on anomalous regions, as evidenced by attention visualizations 
and per-layer logit analysis in Sec.~\ref{sec:attn} and Appendix~\ref{app:layer}.

\begin{table}[t]
    \centering
    \begin{minipage}{0.405\textwidth}
        \centering
        \caption{Ablation study on prompts for the expert model. We evaluate visual and textual prompts and their combination.}
        \resizebox{\linewidth}{!}{
            \begin{tabular}{lcccc}
            \toprule[0.5mm]
            \multirow{2}{*}{\textbf{Setting}} & \multicolumn{2}{c}{MVTec-AD} & \multicolumn{2}{c}{VisA} \\
            \cmidrule(lr){2-3} \cmidrule(lr){4-5}
            & Accuracy$^*$ & F1 & Accuracy$^*$ & F1 \\ \midrule
            Qwen2.5-VL-7B             & 85.9 & 88.3 & 76.1 & 69.4 \\
            + Visual prompt$^\ast$ (full samples) & 75.3 & 89.7 & 68.3 & 76.5 \\
            + Visual prompt          & 78.9 & 87.1 & 68.8 & 76.3 \\
            + Textual prompt          & 82.4 & 99.5 & 65.4 & 77.3 \\
            \rowcolor{gray!20}
            \textbf{+ Visual-Textual} & \textbf{93.6} & \textbf{96.0} & \textbf{86.9} & \textbf{83.7} \\
            \bottomrule[0.5mm]
            \end{tabular}
        }
        \label{tab:prompt_design}
    \end{minipage}
    \hfill
    \begin{minipage}{0.585\textwidth}
        \centering
        \caption{Ablation study on CAAS activation range and target tokens. We compare confidence-aware against full activation, and query image against visual prompt tokens.}
        \resizebox{\linewidth}{!}{
            \begin{tabular}{llcccc}
            \toprule[0.5mm]
            \textbf{Ablation} & \textbf{Setting} & Accuracy$^*$ & Precision & Recall & F1 \\ \midrule

            \multirow{3}{*}{Target tokens}
                & w/o CAAS                        & 93.6 & 96.6 & 95.7 & 96.0 \\
                & CAAS (on query image)   & 94.0 & 96.7 & 95.7 & 96.1 \\
                & \cellcolor{gray!20}CAAS (on visual prompt)  & \cellcolor{gray!20}94.4 & \cellcolor{gray!20}97.5 & \cellcolor{gray!20}94.9 & \cellcolor{gray!20}96.1 \\ \midrule

            \multirow{2}{*}{Activation range}
                
                & CAAS (full range)       & 94.1 & 97.3 & 94.6 & 95.8 \\ 
                & \cellcolor{gray!20}CAAS (low-confidence region)    & \cellcolor{gray!20}94.4 & \cellcolor{gray!20}97.5 & \cellcolor{gray!20}94.9 & \cellcolor{gray!20}96.1 \\ 
            \bottomrule[0.5mm]
            \end{tabular}
        }
        \label{tab:caas_tokens}
    \end{minipage}
    \vspace{-14pt}
\end{table}

\textbf{Impact of CAAS.} As shown in Tab.~\ref{tab:caas_tokens}, we conduct ablation studies on MVTec-AD along two dimensions: target tokens and activation range, to investigate the optimal 
design of CAAS. Regarding target tokens, enhancing attention toward visual prompt tokens outperforms query image tokens, suggesting that the anomalous regions highlighted by red bounding boxes in the visual prompt provide more discriminative cues for correct prediction than the query image itself.
Regarding activation range, CAAS (full range) applies CAAS on query image tokens 
for normal predictions and on visual prompt tokens for anomalous predictions 
across all samples, yet yields limited improvement over the baseline.
Restricting activation to the 
low-confidence region $s_{img} \in [\tau, s_{max}]$ achieves the best 
performance, demonstrating that confidence-aware activation avoids unnecessary 
interference on high-confidence samples. More detailed analysis and ablation studies are provided in Appendix~\ref{app:alpha}.

\begin{table*}[h]
\centering
\vspace{-4pt}
\caption{Per-class anomaly detection performance on MVTec-AD. \colorbox{yellow!20}{Highlighted} categories represent cases where the expert model exhibits limitations and EAGLE provides consistent improvements.}
\vspace{-5pt}
\label{tab:6}
\resizebox{\textwidth}{!}{
\begin{tabular}{ll ccc ccc ccc ccc}
\toprule[0.4mm]
& \multirow{2}{*}{\textbf{Class}}
& \multicolumn{3}{c}{\textbf{Qwen2.5VL-7B}}
& \multicolumn{3}{c}{\textbf{PatchCore}}
& \multicolumn{3}{c}{\textbf{EAGLE w/o CAAS}}
& \multicolumn{3}{c}{\textbf{EAGLE w/ CAAS}} \\
\cmidrule(lr){3-5} \cmidrule(lr){6-8} \cmidrule(lr){9-11} \cmidrule(lr){12-14}
& 
& \textbf{Precision} & \textbf{Recall} & \textbf{Accuracy$^*$}
& \textbf{Precision} & \textbf{Recall} & \textbf{Accuracy$^*$}
& \textbf{Precision} & \textbf{Recall} & \textbf{Accuracy$^*$}
& \textbf{Precision} & \textbf{Recall} & \textbf{Accuracy$^*$} \\
\midrule
& bottle     & 96.4 & 85.7 & 87.9
            & 98.4 & 100.0 & \underline{97.5} 
            & 95.5 & 100.0 & \textbf{99.2 }
            & 95.5 & 100.0 & \textbf{99.2} \\
            
& cable      & 97.5 & 84.8 & 90.7 
            & 98.9 & 93.5 & \textbf{95.9}
            & 96.1 & 90.2 & 92.9
            & 96.2 & 93.9 & \underline{94.7} \\
            
& capsule    & 96.0 & 89.0 & 85.8 
            & 97.1 & 92.7 & \underline{89.8} 
            & 96.4 & 92.0 & \textbf{92.6}
            & 96.4 & 92.0 & \textbf{92.6}\\

& hazelnut   & 88.5 & 98.6 & 88.0 
            & 100.0 & 100.0 & \textbf{100.0 }
            & 100.0 & 98.6 & \underline{99.3 }
            & 100.0 & 100.0 & \textbf{100.0} \\

& metal\_nut & 98.4 & 87.1 & 91.3 
            & 97.2 & 98.6 & \textbf{94.7 }
             & 97.2 & 98.6 & \textbf{94.7}
            & 97.2 & 98.6 & \textbf{94.7} \\
            
& pill       & 99.1 & 80.3 & \textbf{80.2 }
            & 99.0 & 78.8 &  \underline{79.4}
           & 99.0 & 78.8 &  \underline{79.4}
            & 99.0 & 77.3 & 78.6 \\
            
& screw      & 95.7 & 55.5 & 74.1 
            & 97.4 & 94.1 & \textbf{93.4}
            & 97.3 & 92.4 & \underline{92.6}
            & 98.1 & 86.6 & 90.8 \\

& toothbrush & 79.0 & 100.0 & 66.7 
            & 100.0 & 100.0 & \textbf{100.0 }
           & 100.0 & 100.0 & \textbf{100.0 }
            & 100.0 & 100.0 & \textbf{100.0} \\
            
& transistor & 82.8 & 60.0 & 75.8 
            & 100.0 & 97.5 & \textbf{98.8}
            & 100.0 & 97.5 &\textbf{98.8}
            & 100.0 & 92.5 & \underline{96.2} \\

& zipper     & 95.7 & 56.3 & 73.5 
            & 96.7 & 99.2 & \underline{93.3}
            & 95.6 & 95.6 & \textbf{93.9}
            & 95.6 & 95.6 & \textbf{93.9} \\

& tile       & 100.0 & 83.3 & 91.7 
            & 100.0 & 96.4 & \textbf{98.2 }
          & 95.8 & 100.0 & \underline{96.9}
            &95.8 & 100.0 & \underline{96.9} \\
     
\rowcolor{yellow!20}& wood        & 98.4 & 100.0 & \textbf{97.4 }
            & 93.7 & 98.3 & 88.6 
             &93.5 & 96.7 & 87.8
            & 96.7 & 96.7 & \underline{93.1} \\

\rowcolor{yellow!20}& carpet   & 98.8 & 89.9 & \textbf{93.2}
            & 88.8 & 97.8 & 79.2
           & 92.5 & 96.6 & 85.8
            & 94.4 & 94.4 &\underline{ 88.3 }\\
            
\rowcolor{yellow!20}& grid        & 100.0 & 91.2 & 95.6 
            & 96.6 & 98.2 & 93.9 
            & 95.7 & 97.8 & \underline{95.7} 
            & 97.7 & 95.6 & \textbf{96.2 }\\

\rowcolor{yellow!20}& leather    & 98.9 & 96.7 & \underline{96.8 }
            & 86.8 & 100.0 & 78.1 
            & 93.7 & 100.0 & 95.0  
            & 100.0 & 100.0 & \textbf{100.0}\\

\rowcolor{gray!20}
& \textbf{Average}
& 95.0 & 83.9 & 85.9
& 96.7 & 96.3 & 92.1
& 96.6 & 95.7 & \underline{93.6}
& 97.5 & 94.9 & \textbf{94.4}\\
\bottomrule[0.4mm]
\end{tabular}
}
 \vspace{-10pt}
\end{table*}
\subsection{Further Analysis}
\label{sec:attn}
\textbf{When and Why EAGLE Works?} We further investigate the reasons for its success and identify the conditions under which it may underperform. Following the metric 
definitions in Appendix~\ref{app:metrics}, Precision measures the 
proportion of truly anomalous samples among all samples predicted as 
anomalous, while Recall measures the proportion of true anomalies that 
are correctly detected. 

As shown in Tab.~\ref{tab:mvtec_visa} and Tab.~\ref{tab:6}, standalone MLLMs tend to 
yield high Precision but low Recall, suggesting a bias toward classifying 
anomalous samples as normal, i.e., a high false negative (FN) rate. Incorporating 
the expert model with informative prompts alleviates this issue in Tab.~\ref{tab:6}.

\begin{wrapfigure}{r}{0.5\textwidth} 
    \centering
    \vspace{-10pt}
    \includegraphics[width=\linewidth]{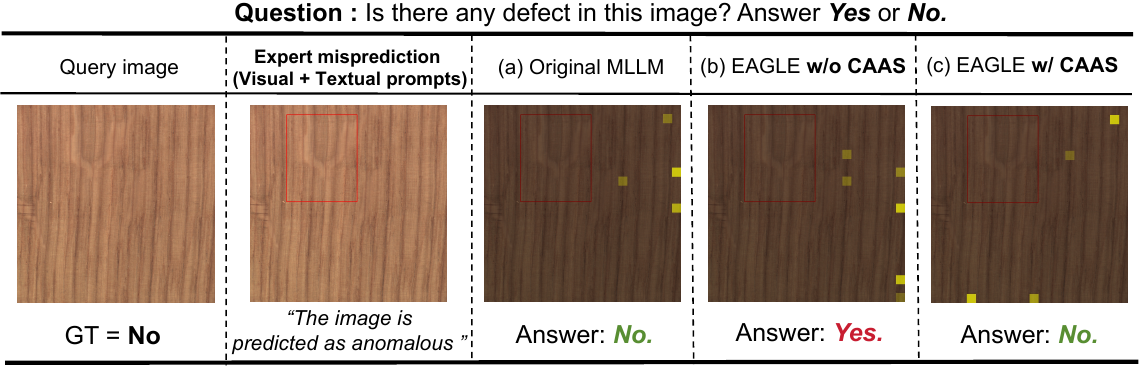}
    \caption{Case of expert model misprediction.}
    \label{fig:7}
\vspace{-10pt}
\end{wrapfigure}
However, the expert model exhibits inherent limitations on certain categories, 
as evidenced in Tab.~\ref{tab:6}, where representative categories are highlighted for detailed analysis. On leather, PatchCore achieves only 86.8\% 
Precision, i.e., a relatively high false positive (FP) rate, as its patch-level 
local feature comparisons are prone to misidentifying normal textural variations 
as defects. In contrast, MLLMs leverage global semantic understanding to 
overlook minor local perturbations. This behavior 
is further illustrated on the wood category through attention visualizations in 
Fig.~\ref{fig:7}, where the standalone MLLM correctly identifies the sample as 
normal, with attention diffusely distributed rather than focused on any specific 
region~(a). However, the expert model mistakenly flags the normal texture as 
anomalous, introducing erroneous prompts that mislead the MLLM into an incorrect 
prediction~(b). CAAS effectively corrects this by redirecting attention toward 
visual tokens, recovering the correct prediction~(c). Such corrections are consistently observed across texture categories, improving Precision on leather 
from 86.8\% to 100.0\%, and on carpet, grid, and wood from 88.8\%, 96.6\%,  
and 93.7\% to 94.4\%, 97.7\%, and 96.7\%, respectively. 

In contrast, when the MLLM exhibits low recall while the expert model maintains 
high precision and recall, EAGLE tends to rely more on the expert model, 
leading to improved overall performance, as observed in categories such as hazelnut and metal\_nut in Tab.~\ref{tab:6}. 
Moreover, when the expert model itself suffers from severe missed detections, 
CAAS is unable to effectively compensate. For instance, on the capsules category of VisA 
in Tab.~\ref{tab:per-class}, PatchCore achieves only 32\% Recall, and the standalone 
MLLM similarly yields 28\%, indicating that neither component can effectively 
detect anomalies, and consequently EAGLE fails to achieve meaningful improvement.


\begin{figure*}[h]    
  \centering    
  \subfloat[Correct vs. incorrect predictions.]   
  {
  \label{fig:subfig1}\includegraphics[width=0.45\textwidth]{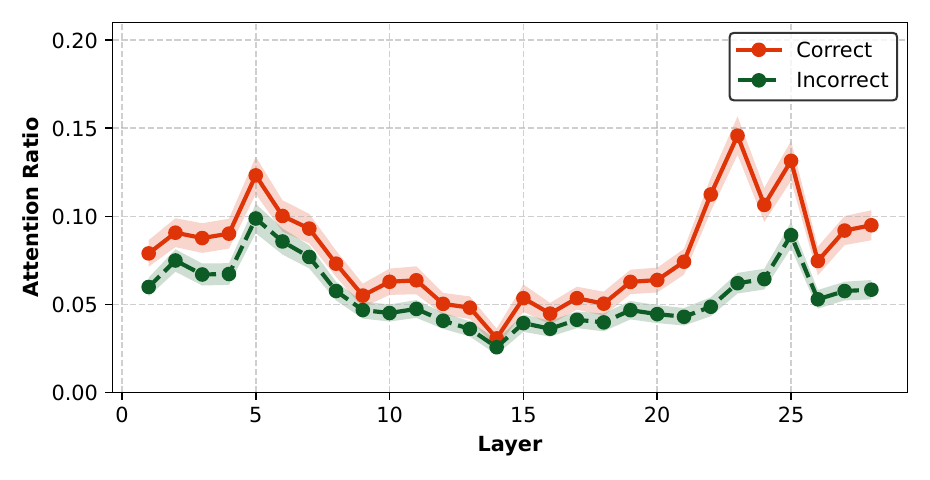}
  }
  \subfloat[EAGLE vs. baseline MLLM.]
  {
    \label{fig:subfig2}\includegraphics[width=0.45\textwidth]{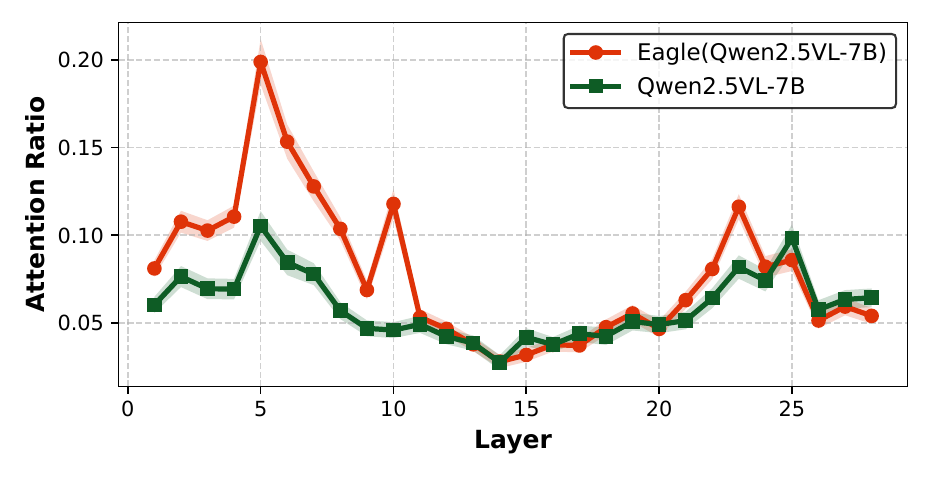}
  }
  \caption{Attention ratios to ground-truth anomalous regions across Transformer layers. Higher attention ratios indicate stronger focus on anomalies. Shaded regions denote standard deviation.}   
  \label{fig:attention}      
  \vspace{-5pt}
\end{figure*}
\textbf{Attention Alignment and Detection Accuracy.}
To investigate whether incorrect predictions are associated with improper attention behaviors, we analyze the visual attention allocated to ground-truth anomalous regions across Transformer layers, focusing on anomalous samples in the test set. Specifically, we define an \emph{attention ratio} to quantify the degree to which the model focuses on true defect regions as $\text{AR}^{(l)} = \frac{\sum_{p \in \mathcal{G}} A^{(l)}_p}{\sum_{p \in \mathcal{F}} A^{(l)}_p}$, 
where $A^{(l)}_p$ denotes the attention weight of the $p$-th image token at the $l$-th Transformer layer, $\mathcal{G}$ represents the ground-truth anomalous region tokens, and $\mathcal{F}$ denotes the set of foreground (object) image tokens instead of all image tokens. 
This normalization is introduced to mitigate the bias that may arise from general foreground attention, ensuring that the metric reflects attention alignment with anomaly regions rather than overall object regions.


\begin{wrapfigure}{r}{0.35\textwidth} 
  \centering
   \vspace{-10pt}
  \includegraphics[width=\linewidth]{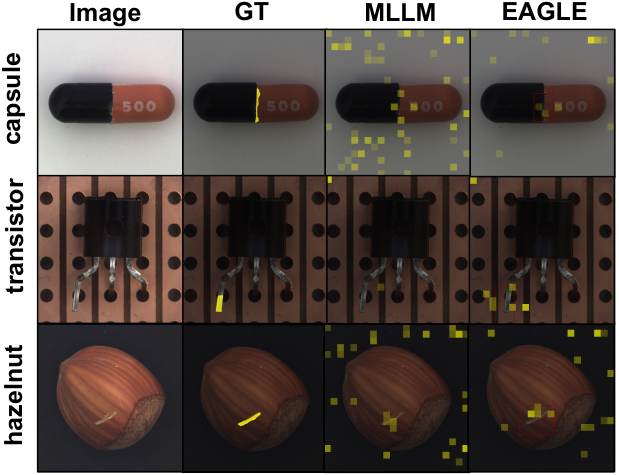}
  \vspace{-10pt}
  \caption{Attention map visualizations from Qwen2.5-VL-7B. }
  \label{fig:10-vis}
\vspace{-15pt}
\end{wrapfigure}

We conduct two experiments to analyze attention ratios across Transformer layers. First, we compare correctly and incorrectly predicted anomalous samples from the baseline MLLM on MVTec-AD. As shown in Fig.~\ref{fig:attention} (a), correct predictions exhibit consistently higher attention ratios, indicating a strong correlation between detection accuracy and attention to ground-truth anomalous regions. Second, we compare the baseline MLLM with EAGLE on all anomalous samples. Fig.~\ref{fig:attention} (b) shows that EAGLE assigns higher attention to anomalous regions across most layers, demonstrating that dual-modality prompts effectively guide the MLLM. Similar trends hold across other MLLMs in Appendix~\ref{app:attention}. To provide qualitative evidence, we visualize attention maps in Fig.~\ref{fig:10-vis}, where our framework shows more focused attention on defect regions, improving visual evidence utilization.

\section{Conclusion}
\vspace{-5pt}

In this paper, we demonstrate that MLLMs can achieve strong industrial anomaly detection performance without any fine-tuning. By incorporating expert-guided prompting through TGPS and CAAS, EAGLE improves detection accuracy and recall across multiple MLLM backbones without any parameter learning. These results suggest that structured expert guidance can serve as a practical alternative to model fine-tuning for industrial anomaly detection. In addition, our analysis reveals a clear relationship between the distribution of visual attention in MLLMs and prediction correctness. We believe this observation opens up promising directions for future work on enhancing anomaly understanding and reasoning capabilities in MLLMs.

\section*{References}

{
    \small
    \bibliographystyle{plain}
    \bibliography{neurips_2025}
}

\newpage

\appendix
\section*{Appendix}

\section{Preliminary on MLLMs}
\label{app:1}
\subsubsection{Multimodal Large Language Models}
In MLLMs, visual tokens extracted by a vision encoder and a projector are concatenated with text tokens, forming a unified input sequence $\mathbf{X}=[\mathbf{X}_v; \mathbf{X}_t]$, which is then fed into the language model. The language model is composed of multiple Transformer layers, among which a core component is the multi-head attention (MHA) mechanism.
\begin{equation}
    x_i^{l} = x_i^{l-1} + \sum_{h=1}^{H} \mathrm{MHA}^{l,h}(x_i^{l-1}),
\end{equation}
\begin{equation}
    \mathrm{MHA}^{l,h}(x_i^{l-1}) = \sum_{j \le i} A_{i,j}^{l,h} \, x_j^{l-1}W_V^{l,h},
\label{eq:7}
\end{equation}
where $x_i^{l-1}$ and $x_i^{l}$ represent the input and output of the $i$-th token of the MHA at layer $l$, respectively, and $H$ is the number of attention heads. Eq.\ref{eq:7} represents the individual input $x_i^{l-1}$ and its previous tokens attention contributions $X^{l-1}_{\leq i} = \{x_1^{l-1},...,x_i^{l-1}\}$. Here, $A_{i,j}^{l,h}$ denotes the attention weight from $x_j^{l-1}$ to $x_i^{l-1}$, which can be interpreted as the degree to which $x_i^{l-1}$ attends to $x_j^{l-1}$. $W_V$ represents the value matrix.

Finally, for a sequence of length L, the probability of generating the answer $\mathbf{X}_a$ is computed in an autoregressive manner: 
\begin{equation}
    p(\mathbf{X}_a \mid \mathbf{X}_v, \mathbf{X}_t)
    = \prod_{i=1}^{L} p\!\left(x_i \mid \mathbf{X}_v, \mathbf{X}_{t,<i}, \mathbf{X}_{a,<i}\right),
\end{equation}
where $\mathbf{X}_{t,<i}$ and $\mathbf{X}_{a,<i}$ denote the text tokens and previously generated answer tokens before position $i$, respectively.

From this formulation, it follows that attention weights play a critical role in controlling how visual and textual information is aggregated, and modifying these weights can influence the model’s final prediction.

\subsubsection{Visual Attention Distribution in MLLMs}
\label{app:2}
\begin{wrapfigure}{r}{0.5\textwidth} 
\vspace{1pt}
    \centering
    \vspace{-15pt}
    \includegraphics[width=\linewidth]{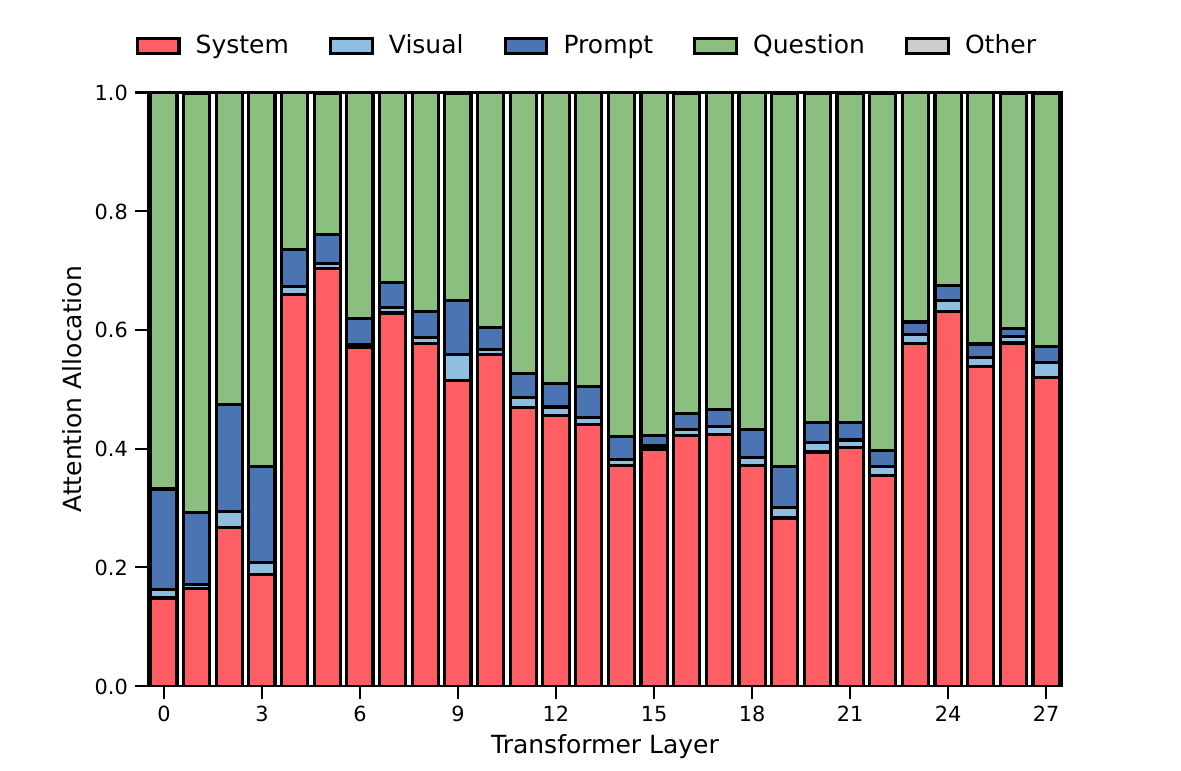}
    \caption{Attention allocation between visual and textual tokens across different layers of the Qwen2.5VL-7B.
}
   \label{fig:alloc}
   \vspace{-.3cm}
\end{wrapfigure}

Recent studies have revealed a systematic attention imbalance in MLLMs, where textual tokens receive disproportionately higher attention than visual tokens. \cite{chen2025spatial} quantitatively demonstrate this imbalance: despite image tokens comprising approximately 90\% of the input sequence, they receive only around 10\% of the total attention weight, indicating that MLLMs fundamentally underutilize visual information. As a result, textual priors tend to dominate over visual evidence, particularly in vision-centric tasks. \cite{yin2025clearsight} further analyze this phenomenon from the perspective of modality fusion, showing that in the middle Transformer layers, where cross-modal fusion is most critical, attention allocated to visual features is significantly lower than that to textual tokens, causing the output distribution to be skewed toward language priors and leading to object hallucination. \cite{kang2025see} additionally identify that the limited visual attention is further wasted on semantically irrelevant background tokens (visual attention sinks), leaving even less effective attention for meaningful visual content.

Following~\cite{yin2025clearsight}, we analyze the attention allocation across different Transformer layers of Qwen2.5VL-7B. As illustrated in Fig.~\ref{fig:alloc}, visual tokens consistently receive a substantially lower proportion of attention compared to textual tokens across nearly all layers, particularly in the middle layers where modality fusion is most critical. This observation is consistent with the findings reported in~\cite{yin2025clearsight}, confirming that MLLMs tend to underutilize visual information during inference, leading to predictions biased toward language priors. This imbalance provides the motivation for our proposed CAAS module, which explicitly amplifies attention toward visual tokens to enhance the model's focus on anomalous regions.

\section{Details of Evaluation Metrics}
\label{app:metrics}

For the anomaly detection task, we adopt a binary classification evaluation 
protocol, consistent with the Anomaly Discrimination task defined in the 
MMAD benchmark~\cite{jiang2024mmad}. We treat the anomalous class as the 
positive class and the normal class as the negative class. Model performance 
is assessed using the following metrics:

\textbf{Precision:} The proportion of correctly identified 
anomalous samples among all samples predicted as anomalous:
\begin{equation}
\text{Precision} = \frac{TP}{TP + FP}
\end{equation}

\textbf{Recall:} The proportion of correctly identified anomalous 
samples among all true anomalous samples:
\begin{equation}
\text{Recall} = \frac{TP}{TP + FN}
\end{equation}

\textbf{F1-score:} The harmonic mean of Precision and Recall:
\begin{equation}
\text{F1} = \frac{2 \times \text{Precision} \times \text{Recall}}
{\text{Precision} + \text{Recall}}
\end{equation}

\textbf{Accuracy$^*$ (Balanced):} Since the number of normal samples 
typically far exceeds that of anomalous samples in industrial anomaly 
detection benchmarks, standard accuracy tends to be biased toward the 
majority class. Therefore, we report Balanced Accuracy, defined as the 
arithmetic mean of normal-class accuracy (Specificity) and anomalous-class 
accuracy (Recall), following the same computation used in~\cite{jiang2024mmad}:
\begin{equation}
\text{Accuracy$^*$} = \frac{1}{2}
\left(\frac{TN}{TN + FP} + \frac{TP}{TP + FN}\right)
\end{equation}
where $TP$, $TN$, $FP$, and $FN$ denote true positives, true negatives, 
false positives, and false negatives, respectively. The final reported 
score is the mean Balanced Accuracy averaged across all classes.

\textbf{Accuracy.} Accuracy measures the overall classification performance by computing the ratio of correctly classified samples (both normal and anomalous) to the total number of samples:
\begin{equation}
\mathrm{Accuracy} = \frac{TP + TN}{TP + TN + FP + FN}
\end{equation}

\section{Implementation Detail} 
\label{app:implementation detail}
We adopt PatchCore\cite{roth2022towards} with WideResNet-50 as the backbone. Images are processed at a resolution of $256 \times 256$ and center-cropped to $224 \times 224$ for PatchCore, while visual prompts are restored to the original image resolution before being fed into the MLLMs. For the experiments reported in Tab.~\ref{tab:mvtec_visa} and ~\ref{tab:comparison_mllm}, we set the parameter $\alpha = 0.6$. Further Automatic threshold selection is discussed in Appendix~\ref{app:threshold}. For original MLLMs, we follow the 1-shot+ setting proposed in~\cite{jiang2024mmad}, where the most similar normal image is retrieved from the training set and used as a template to provide additional visual context during inference. All experiments were conducted on a single RTX 3090 24GB GPU.

\begin{table}[h]
\centering
\caption{Comparison of average performance across tasks on MMAD.}
\label{tab:app mmad}
\resizebox{1\linewidth}{!}{
\begin{tabular}{lcccccccc}
\toprule[0.5mm]
\textbf{Dataset} 
& \textbf{Anomaly Detection} 
& \textbf{Object Classification} 
& \textbf{Object Analysis} 
& \textbf{Defect Classification} 
& \textbf{Defect Localization} 
& \textbf{Defect Description} 
& \textbf{Defect Analysis} 
& \textbf{Average} \\
\midrule 
MVTec-AD 
& 93.45 & 83.72 & 92.30 & 67.31 & 75.14 & 81.11 & 91.18 & 83.46 \\
VisA 
& 88.52 & 80.07 & 82.31 & 53.40 & 54.13 & 67.44 & 76.74 & 71.80 \\

MVTec-LOCO 
& 52.57 & 82.04 & 80.44 & 41.52 & 43.54 & 65.03 & 74.56 & 62.81 \\

GoodsAD 
& 55.36 & 89.68 & 77.13 & 50.93 & 48.19 & 58.55 & 75.25 & 65.01 \\
\rowcolor{gray!20}Average
& 72.48 & 83.88 & 83.05 & 53.29 & 55.25 & 68.03 & 79.43 & 70.77 \\
\bottomrule[0.5mm]
\end{tabular}
}
\end{table}

\section{Additional Analysis and Results}

\begin{table*}[!b]
\centering
\caption{Per-class performance across six datasets. The decision threshold is estimated via the $k$-sigma method within TGPS, demonstrating consistent results with Tab.~\ref{tab:6}.}
\label{tab:per-class}
\resizebox{\textwidth}{!}{
\begin{tabular}{llcccccccc cccc}
\toprule[0.5mm]
\multirow{2}{*}{\textbf{Dataset}} & \multirow{2}{*}{\textbf{Class}}
& \multicolumn{4}{c}{\textbf{PatchCore}}
& \multicolumn{4}{c}{\textbf{Eagle}} 
& \multicolumn{4}{c}{\textbf{Qwen2.5VL-7B}}\\
\cmidrule(lr){3-6} \cmidrule(lr){7-10} \cmidrule(lr){11-14}
&
& \textbf{Precision} & \textbf{Recall} & \textbf{F1} & \textbf{Accuracy}
& \textbf{Precision} & \textbf{Recall} & \textbf{F1} & \textbf{Accuracy}
& \textbf{Precision} & \textbf{Recall} & \textbf{F1} & \textbf{Accuracy} \\
\midrule
\midrule
\multirow{16}{*}{\textbf{MVTec-AD}}
& bottle     & 98.44 & 100.00 & 99.21 & \textbf{97.50}
              & 98.44 & 100.00 & 99.21 & \textbf{97.50}
              & 96.43 & 85.71 & 90.76 & 87.86 \\

& cable      & 100.00 & 92.39 & 96.05 & \textbf{96.20} 
              & 100.00 & 91.30 & 95.45 &\underline{95.65}
              & 97.50 & 84.78 & 90.70 & 90.67 \\

& capsule    & 97.22 & 96.33 & 96.77 &\underline{91.64}
              & 98.13 & 96.33 & 97.22 & \textbf{93.82}
              & 96.04 & 88.99 & 92.38 & 85.80 \\

& carpet     & 90.63 & 97.75 & 94.05 & 82.80
              & 94.44 & 95.51 & 94.97 & \underline{88.82}
              & 98.77 & 89.89 & 94.12 & \textbf{93.16} \\

& grid        & 98.25 & 98.25 & 98.25 & \underline{96.49}
              & 100.00 & 98.25 & 99.12 & \textbf{99.12}
              & 100.00 & 91.23 & 95.41 & 95.61 \\

& hazelnut   & 100.00 & 100.00 & 100.00 & \textbf{100.00} 
              &100.00 & 100.00 & 100.00 & \textbf{100.00}
              & 88.46 & 98.57 & 93.24 & 88.04 \\

& leather    & 89.32 & 100.00 & 94.36 & 82.81
              & 100.00 & 100.00 & 100.00 &\textbf{100.00}
              & 98.89 & 96.74 & 97.80 & \underline{96.81}\\

& metal\_nut & 97.18 & 98.57 & 97.87 & \textbf{94.74}
              & 97.18 &98.57 & 97.87 & \textbf{94.74}
             &98.39 & 87.14 & 92.42 & 91.30 \\

& pill      & 99.00 & 75.00 & 85.34 & \underline{77.50}
              & 99.00 & 75.00 & 85.34 & \underline{77.50}
              & 99.07 & 80.30 &88.70 & \textbf{80.15} \\

& screw      & 98.15 & 89.08 & 93.39 & \underline{92.10}
              & 99.06 & 88.24 & 93.33 & \textbf{92.90}
              & 95.65 & 55.46 & 70.21 & 74.07 \\

& tile       & 100.00 & 96.43 & 98.18 & \textbf{98.21}
              & 100.00 & 96.43 & 98.18 &\textbf{98.21}
              & 100.00 & 83.33 & 90.91 & 91.67 \\

& toothbrush & 100.00 & 100.00 & 100.00 & \textbf{100.00} 
              & 100.00 & 100.00 & 100.00 & \textbf{100.00}
              & 78.95 & 100.00 & 88.24 & 66.67 \\

& transistor & 100.00 & 95.00 & 97.44 & \textbf{97.50} 
              & 100.00 & 95.00 & 97.44 & \textbf{97.50}
              & 82.76 & 60.00 & 69.57 & 75.83 \\

& wood       & 93.65 & 98.33 & 95.93 & 88.64
              & 95.08 & 96.67 & 95.87 & \underline{90.44}
              & 98.36 & 100.00 & 99.17 & \textbf{97.37} \\

& zipper     & 96.72 & 99.16 & 97.93 & \textbf{93.33}
              & 96.46 & 99.09 & 97.76 & \underline{93.30}
              &95.71 & 56.30 & 70.90 & 73.46 \\

& \cellcolor{gray!20}\textbf{Average} 
& \cellcolor{gray!20}97.24 & \cellcolor{gray!20}95.75 & \cellcolor{gray!20}96.32 & \cellcolor{gray!20}\underline{92.63}
& \cellcolor{gray!20}98.52 & \cellcolor{gray!20}95.36 & \cellcolor{gray!20}96.78 & \cellcolor{gray!20}\textbf{94.63}
&\cellcolor{gray!20} 95.00 & \cellcolor{gray!20}83.90 & \cellcolor{gray!20}88.30 & \cellcolor{gray!20}85.90 \\
\midrule
\midrule
\multirow{13}{*}{\textbf{VisA}}
& candle       
& 96.04 & 97.00 & 96.52 & \textbf{96.50}
& 96.04 & 97.00 & 96.52 & \textbf{96.50}
& 82.18 & 83.00 & 82.59 & 82.50  \\

& capsules   
& 96.97 & 32.00 & 48.12 & \textbf{65.17}
& 96.88 & 31.00 & 46.97 & \underline{64.67} 
& 100.00 & 28.00 & 43.75 & 64.00\\

& cashew      
& 98.68 & 75.00 & 85.23 &\textbf{ 86.50 }
& 100.00 & 63.00 & 77.30 & \underline{81.50 } 
& 100.00 & 63.00 & 77.30 & \underline{81.50}\\

& chewinggum  
& 98.95 & 94.95 & 96.91 & \textbf{96.47 }
& 100.00 & 84.85 & 91.80 & \underline{92.42}  
& 94.90 & 93.94 & 94.42 & 91.97\\

& fryum       
& 96.47 & 82.00 & 88.65 & \textbf{88.00}
& 96.43 & 81.00 & 88.04 & \underline{87.50 }
& 94.74 & 72.00 & 81.82 & 82.00 \\

& macaroni1 
& 94.12 & 80.00 & 86.49 & \textbf{87.50 }
& 94.94 & 75.00 & 83.80 & \underline{85.50 }
& 91.80 & 56.00 & 69.57 & 75.50\\

& macaroni2  
& 82.93 & 34.00 & 48.23 & \textbf{63.50 }
& 76.60 & 36.00 & 48.98 &\underline{ 62.50 }
& 71.79 & 28.00 & 40.29 & 58.50 \\

& pcb1        
& 94.95 & 94.00 & 94.47 &  \textbf{94.45} 
& 94.79 & 91.00 & 92.86 & \underline{92.95 }
& 94.12 & 32.00 & 47.76 & 64.98\\

& pcb2       
& 98.81 & 83.00 & 90.22 &  \textbf{90.99}
& 98.81 & 83.00 & 90.22 &  \textbf{90.99}
& 78.85 & 41.00 & 53.95 & 64.88 \\

& pcb3        
& 94.79 & 91.92 & 93.33 &  \textbf{93.41} 
& 94.79 & 91.92 & 93.33 &  \textbf{93.41 }
& 87.50 & 42.42 & 57.14 & 68.15\\

& pcb4       
& 93.33 & 98.00 & 95.61 &  \textbf{95.53 }
& 93.33 & 98.00 & 95.61 &  \textbf{95.53}
& 80.99 & 98.00 & 88.69 & 87.61\\

& pipe\_fryum 
& 98.98 & 97.00 & 97.98 &  \textbf{97.18}
& 98.97 & 96.00 & 97.46 & \underline{96.68}
& 94.23 & 98.00 & 96.08 & 91.11 \\

& \cellcolor{gray!20}\textbf{Average} 
& \cellcolor{gray!20}95.42 & \cellcolor{gray!20}79.91 & \cellcolor{gray!20}85.15 & \cellcolor{gray!20}\textbf{87.93} 
& \cellcolor{gray!20}95.13 & \cellcolor{gray!20}77.31 & \cellcolor{gray!20}83.57 & \cellcolor{gray!20}\underline{86.68} 
& \cellcolor{gray!20}89.26 & \cellcolor{gray!20}61.28 & \cellcolor{gray!20}69.45 & \cellcolor{gray!20}76.06  \\
\midrule
\midrule

\multirow{6}{*}{\textbf{MVTec-LOCO}}
& breakfast\_box       
& 63.24 & 99.42 & 77.30 & 50.69 
&  70.97 & 25.43 & 37.45 & \underline{53.89}
& 87.04 & 27.17 & 41.41 & \textbf{60.15}\\

& juice\_bottle        
& 71.52 & 100.00 & 83.39 & 50.00 
& 72.41 & 53.39 & 61.46 & \underline{51.16 }
& 98.92 & 38.98 & 55.93 & \textbf{68.96} \\

& pushpins             
& 55.34 & 100.00 & 71.25 & 50.00 
&  56.78 & 39.18 & 46.37 & \underline{51.11}
& 86.21 & 14.62 & 25.00 & \textbf{55.86} \\

& screw\_bag           
& 71.22 & 45.21 & 55.31 & \underline{56.21}  
& 80.65 & 11.42 & 20.00 & 53.25 
& 83.33 & 29.68 & 43.77 & \textbf{59.51}  \\

& splicing\_connectors 
& 62.26 & 100.00 & 76.74 & 50.00 
& 85.11 & 21.28 & 34.04 & \underline{57.65} 
& 69.38 & 57.51 & 62.89 & \textbf{57.82}\\

& \cellcolor{gray!20}\textbf{Average} 
& \cellcolor{gray!20}64.71 &\cellcolor{gray!20}88.93 & \cellcolor{gray!20}72.80 & \cellcolor{gray!20}51.38
& \cellcolor{gray!20}73.18 & \cellcolor{gray!20}30.14 & \cellcolor{gray!20}39.86 & \cellcolor{gray!20}\underline{53.41}
& \cellcolor{gray!20}84.98 & \cellcolor{gray!20}33.59 & \cellcolor{gray!20}45.80 & \cellcolor{gray!20}\textbf{60.46}  \\
\midrule
\midrule

\multirow{4}{*}{\textbf{BTAD}}
& 01 
& 100.00 & 89.80 & 94.62 & \textbf{94.90} 
& 100.00 & 89.80 & 94.62 & \textbf{94.90}
& 100.00 & 30.61 & 46.88 & 65.31 \\

& 02 
& 99.01  & 50.00 & 66.45 & \textbf{73.33}
& 99.01 & 50.00 & 66.45 & \textbf{73.33 }
& 100.00 & 42.00 & 59.15 & 71.00 \\

& 03  
& 91.11  & 100.00 & 95.35 &\textbf{ 99.50}
& 91.11 & 100.00 & 95.35 & \textbf{99.50}
& 76.32  & 70.73 & 73.42 & 84.24 \\

& \cellcolor{gray!20}\textbf{Average}
& \cellcolor{gray!20}96.71 & \cellcolor{gray!20}79.93 & \cellcolor{gray!20}85.47 & \cellcolor{gray!20}\textbf{89.24} 
&\cellcolor{gray!20}96.71 & \cellcolor{gray!20}79.93 & \cellcolor{gray!20}85.47 & \cellcolor{gray!20}\textbf{89.24} 
& \cellcolor{gray!20}92.11 &\cellcolor{gray!20} 47.81 & \cellcolor{gray!20}59.82 &\cellcolor{gray!20}73.52  \\
\midrule
\midrule

\multirow{7}{*}{\textbf{MPDD}}
& bracket\_black
& 93.10 & 57.45 & 71.05 & \textbf{75.60} 
& 93.10 & 57.45 & 71.05 & \textbf{75.60 }
& 68.75 & 70.21 & 69.47 & 61.67 \\

& bracket\_brown
& 92.00 & 90.20 & 91.09 & \underline{87.41 }
& 93.75 & 88.24 & 90.91 & \textbf{88.35} 
& 63.27 & 60.78 & 62.00 & 45.78  \\

& bracket\_white
& 65.85 & 90.00 & 76.06 & \textbf{71.67} 
& 65.85 & 90.00 & 76.06 & \textbf{71.67 }
& 55.56 & 33.33 & 41.67 & 53.33  \\

& connector      
& 87.50 & 100.00 & 93.33 & \textbf{96.67 }
& 87.50 & 100.00 & 93.33 & \textbf{96.67} 
& 61.11 & 78.57 & 68.75 & 77.61 \\

& metal\_plate   
& 100.00 & 100.00 & 100.00 & \textbf{100.00} 
& 100.00 & 100.00 & 100.00 & \textbf{100.00 } 
& 84.34 & 98.59 & 90.91 & 74.30 \\

& tubes          
& 100.00 & 62.32 & 76.79 & \textbf{81.16 }
& 100.00 & 62.32 & 76.79 & \textbf{81.16 } 
& 100.00 & 14.49 & 25.32 & 57.25 \\

& \cellcolor{gray!20}\textbf{Average}
& \cellcolor{gray!20}89.74 & \cellcolor{gray!20}83.33 & \cellcolor{gray!20}84.72 & \cellcolor{gray!20}\underline{85.42}
& \cellcolor{gray!20}90.03 & \cellcolor{gray!20}83.00 & \cellcolor{gray!20}84.69 & \cellcolor{gray!20}\textbf{85.57} 
& \cellcolor{gray!20}72.17 & \cellcolor{gray!20}59.33 & \cellcolor{gray!20}59.69 & \cellcolor{gray!20}61.66\\

\midrule
\midrule

\multirow{5}{*}{\textbf{RAD}}
&bolt  
& 95.45 & 88.82 & 92.02 & \underline{84.82} 
& 97.34 & 77.34 & 86.20 & 83.88 
& 96.01 & 94.56 & 95.28 & \textbf{88.38}\\

&ribbon 
& 95.44 & 100.00 & 97.67 & \textbf{90.41}
& 97.41 & 89.76 & 93.43 & \underline{90.09}
& 95.11 & 86.35 & 90.52 & 84.27  \\

&sponge
& 95.25 & 100.00 & 97.57 & \textbf{90.41}
& 97.11 & 83.63 & 89.87 & \underline{87.02}
& 91.98 & 53.02 & 67.27 & 67.61 \\

&tape   
& 96.13 & 100.00 & 98.03 & \underline{91.10 }
& 98.18 & 100.00 & 99.08 & \textbf{95.89}
& 96.13 & 100.00 & 98.03 & 91.10 \\

&  \cellcolor{gray!20}\textbf{Average} 
& \cellcolor{gray!20}95.57 &\cellcolor{gray!20} 97.21 & \cellcolor{gray!20}96.32 & \cellcolor{gray!20}\underline{89.18}
& \cellcolor{gray!20}97.51 & \cellcolor{gray!20}87.68 & \cellcolor{gray!20}92.14 &  \cellcolor{gray!20}\textbf{89.22 }
& \cellcolor{gray!20}94.81 & \cellcolor{gray!20}83.48 & \cellcolor{gray!20}87.77 &\cellcolor{gray!20}82.84\\

\bottomrule[0.5mm]
\end{tabular}
}
\end{table*}

\subsection{Further Analyses and Results on Distribution-Based Thresholding mechanism}
\label{app:addition_results}

\label{app:AS}

\textbf{Conditions for EVT applicability.} 
In the context of industrial anomaly detection, the characteristics of commonly 
used datasets provide strong support for the applicability of EVT. Industrial 
images are typically captured under strictly controlled conditions, including 
fixed illumination, viewpoint, and background, resulting in limited intra-class 
variation among normal samples that can be reasonably approximated as being drawn 
from a common underlying distribution. 

Under this assumption, the image-level anomaly score of a normal sample, defined as the maximum over patch-level anomaly 
scores (Eq.~\ref{max_distance_2}), naturally corresponds to an extreme value 
statistic (Eq.~\ref{eq:evt}). According to EVT, the distribution of such maxima 
can be effectively modeled by an extreme value distribution regardless of the 
specific form of the underlying patch-level score distribution, provided mild 
regularity conditions are satisfied. This property makes EVT particularly 
well-suited for modeling the tail behavior of anomaly scores derived from normal 
training data.

\begin{figure}[htbp]
\centering
\begin{subfigure}{1\columnwidth}
    \centering
    \includegraphics[width=\linewidth]{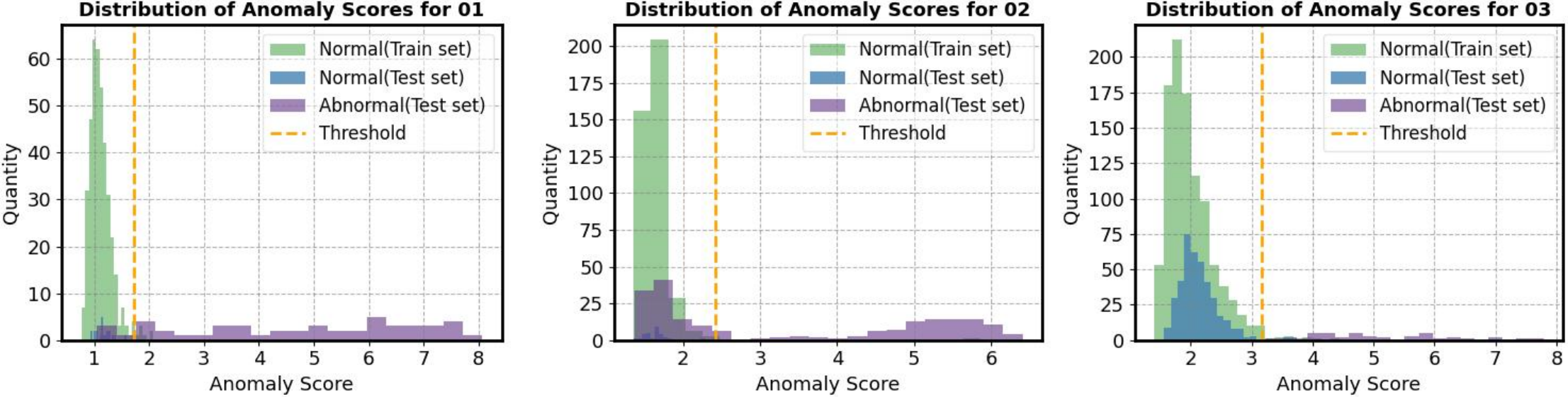}
    \caption{BTAD}
\end{subfigure}
\begin{subfigure}{1\columnwidth}
    \centering
    \includegraphics[width=\linewidth]{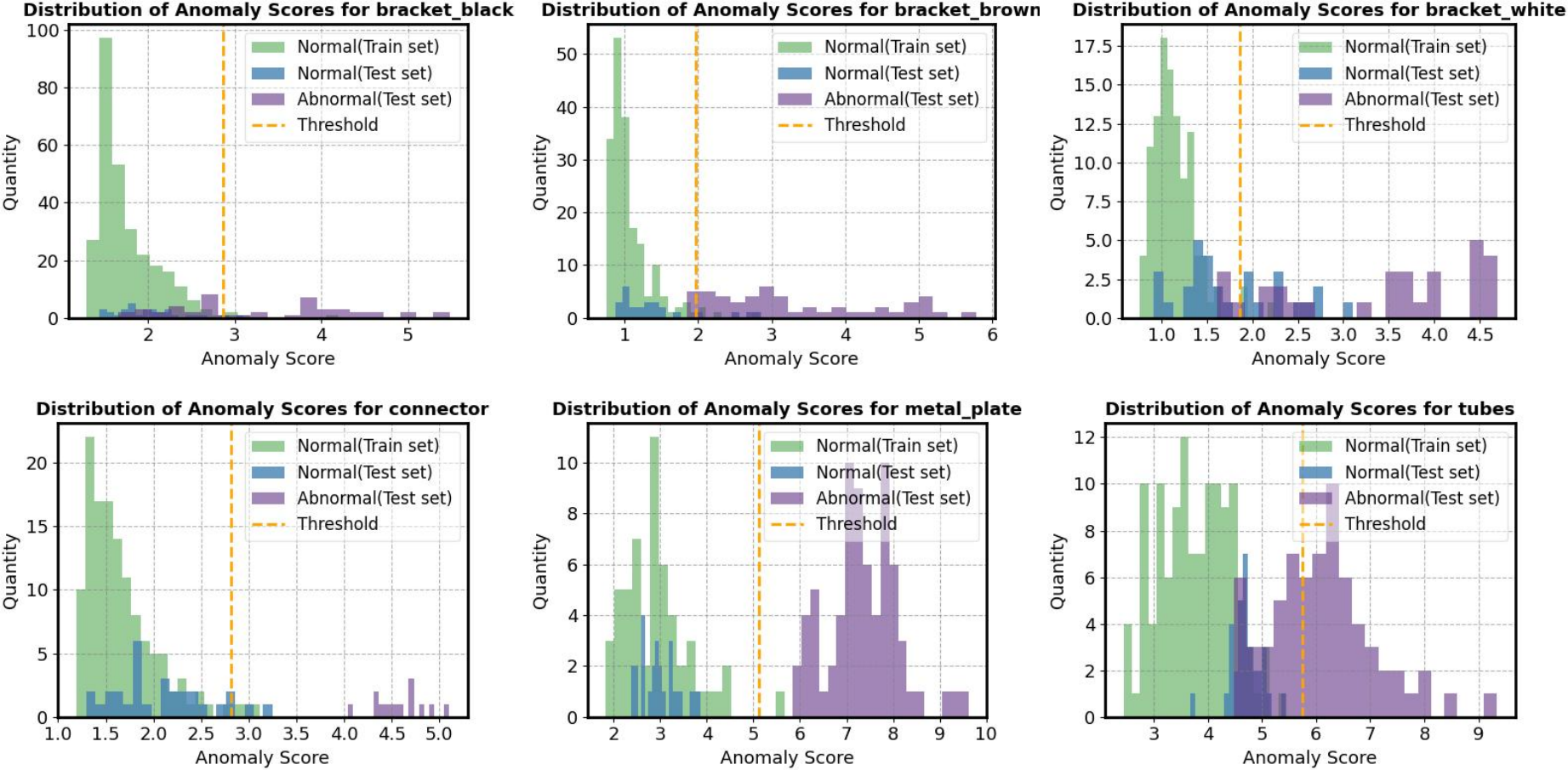}
    \caption{MPDD}
    \label{fig:BTAD-MPDD}
\end{subfigure}
\vspace{0.5em}

\begin{subfigure}{0.48\columnwidth}
    \centering
    \includegraphics[width=\linewidth]{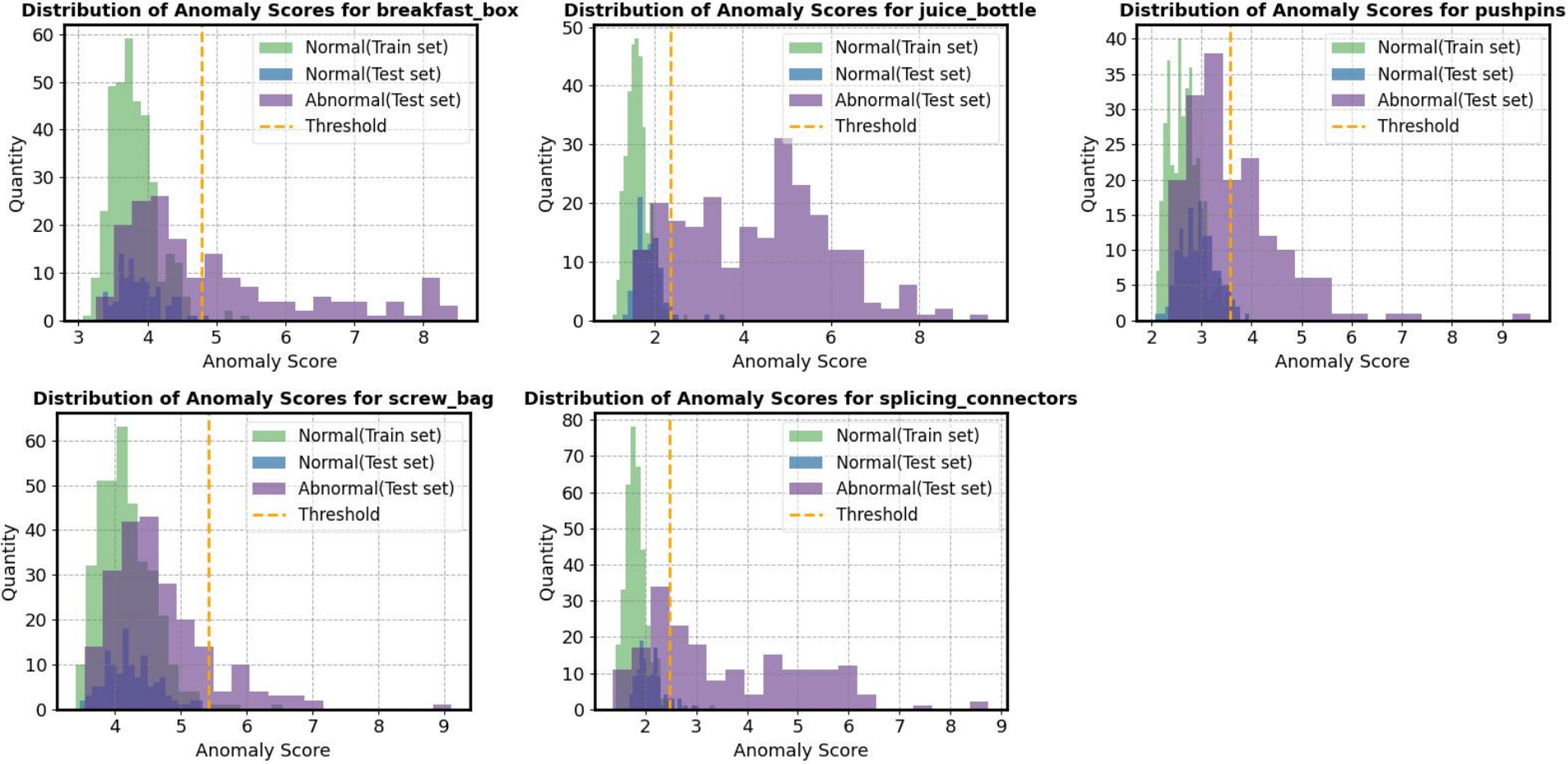}
    \caption{MVTec-LOCO}
\end{subfigure}
\hfill
\begin{subfigure}{0.48\columnwidth}
    \centering
    \includegraphics[width=\linewidth]{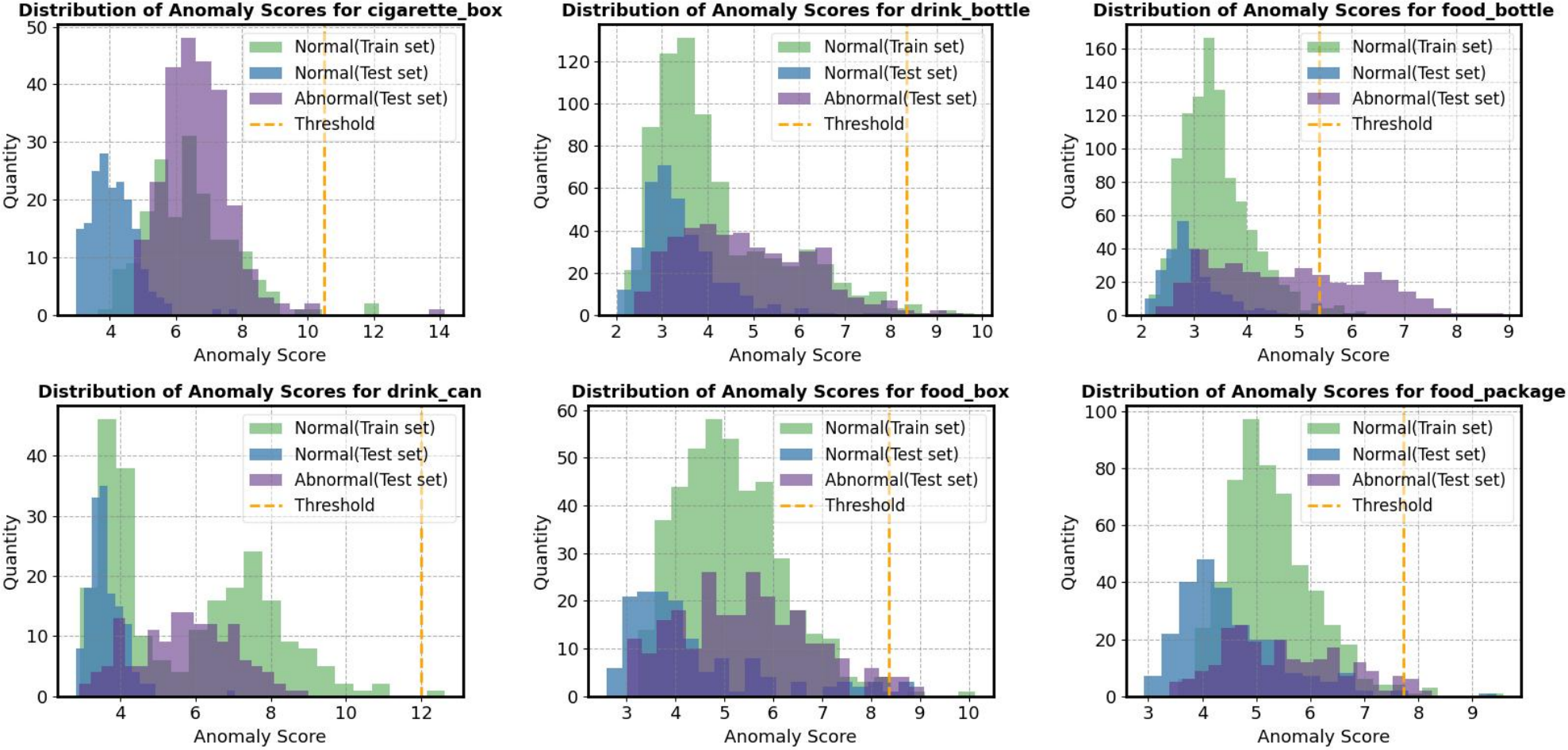}
    \caption{GoodsAD}
\end{subfigure}

\caption{Anomaly score distributions across datasets.}
\label{fig:as_all}
\end{figure}

\begin{figure}
\centering

\begin{minipage}{1\columnwidth}
    \centering
    \includegraphics[width=\linewidth]{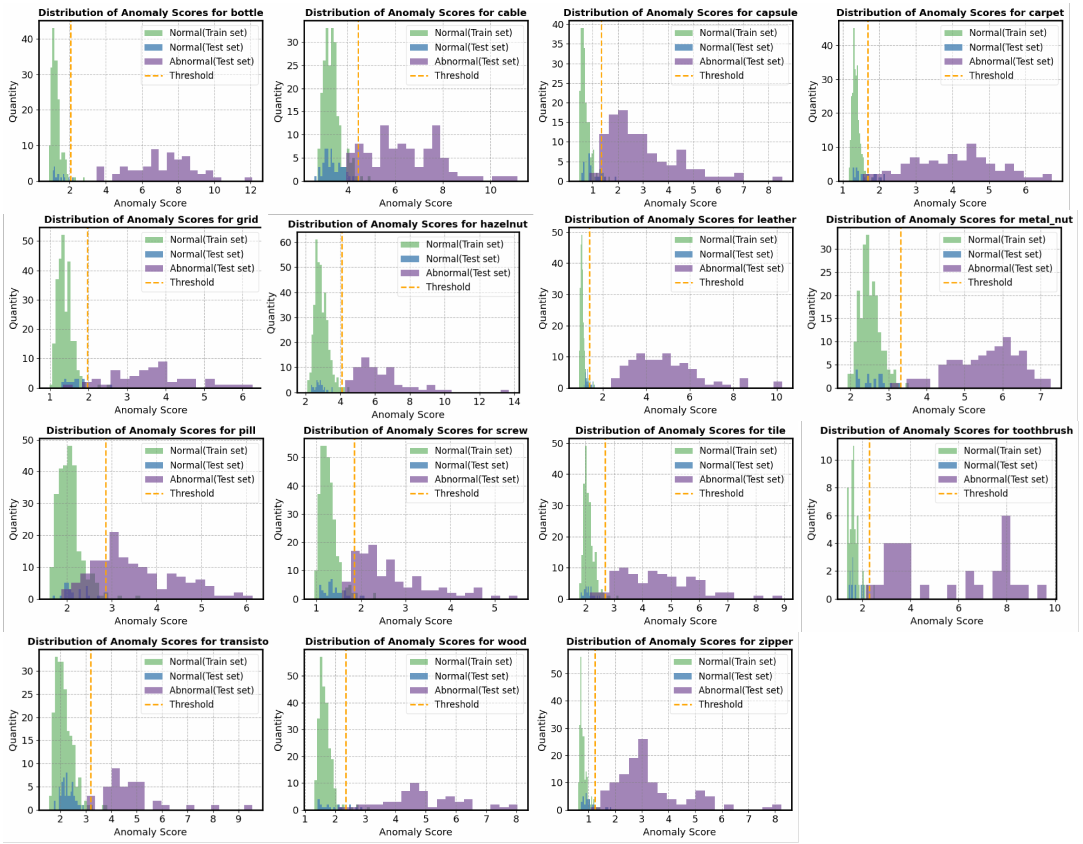}
    \caption{Anomaly score distribution of MVTec-AD.}
    \label{app:as-mvtec}
\end{minipage}

\begin{minipage}{1\columnwidth}
    \centering
    \includegraphics[width=\linewidth]{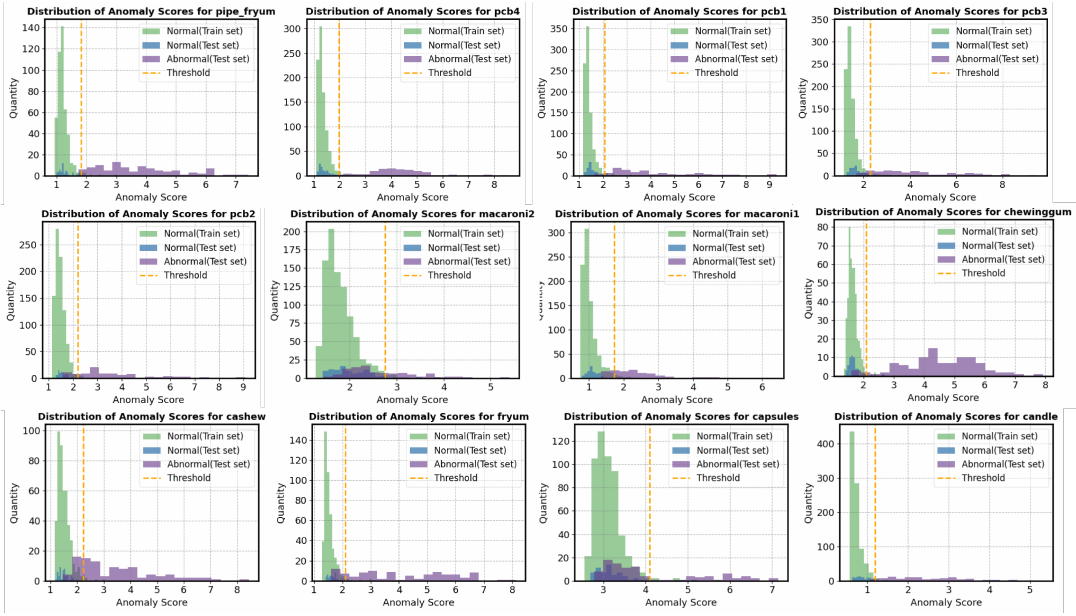}
    \caption{Anomaly score distribution of VisA.}
    \label{app:as-visa}
\end{minipage}

\vspace{0.5em}

\begin{minipage}{1\columnwidth}
    \centering
    \includegraphics[width=\linewidth]{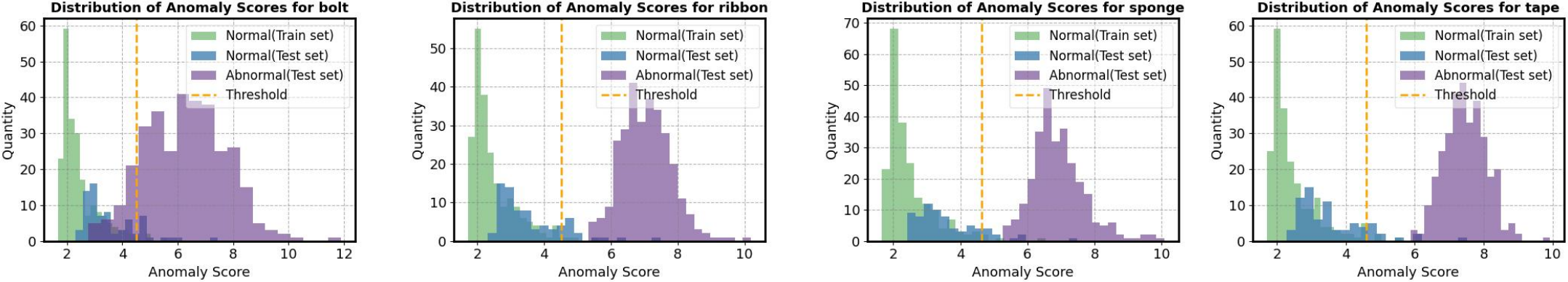}
    \caption{Anomaly score distribution of RAD.}
    \label{app:as-rad}
\end{minipage}
\end{figure}

\textbf{Empirical validation across datasets.} 
Tab.~\ref{tab:per-class} presents per-class performance comparisons between 
PatchCore and EAGLE across six industrial datasets, and Tab.~\ref{tab:app mmad} 
reports the average performance of EAGLE on different tasks of the MMAD benchmark. 
As illustrated in Fig.~\ref{fig:as_all}--\ref{app:as-rad}, the intra-class anomaly 
score distributions of MVTec, VisA, BTAD, RAD, and MPDD are empirically better aligned with EVT-based thresholding, and EAGLE achieves consistently strong performance on these datasets.
In contrast, the intra-class distributions of MVTec-LOCO 
(Tab.~\ref{tab:per-class} and Fig.~\ref{fig:as_all}~(c)) and GoodsAD 
(Tab.~\ref{tab:app mmad} and Fig.~\ref{fig:as_all}~(d)) deviate from the EVT assumption, 
leading to less reliable threshold estimation and consequently degraded detection 
performance, which is consistent with our theoretical analysis.

Notably, the standalone MLLM achieves the best performance on the logical anomaly 
dataset MVTec-LOCO, demonstrating the strong logical reasoning capability of 
MLLMs. Within EAGLE, however, the expert model introduces suboptimal prompts 
on this dataset, slightly degrading performance relative to the standalone MLLM. 
Since our framework treats the MLLM as a plug-and-play component, the expert 
model can be selectively disabled for datasets where it underperforms, providing 
flexibility in practical deployment.


\begin{wraptable}{r}{0.5\textwidth}
    \centering
    \vspace{-12pt}
    \caption{Performance comparison under different thresholding strategies on InternVL3-8B.}
    \vspace{-3pt}
    \resizebox{.92\linewidth}{!}{
            \begin{tabular}{lcccc}
            \toprule[0.5mm]
            \multirow{2}{*}{\textbf{Setting}} & \multicolumn{2}{c}{\textbf{MVTec-AD}} & \multicolumn{2}{c}{\textbf{VisA}} \\ 
            \cmidrule(lr){2-3} \cmidrule(lr){4-5} 
                                        & Accuracy & F1 & Accuracy & F1               \\ \midrule
            K-sigma                     & 93.1 & 95.6 & \textbf{88.3} & \textbf{85.7}           \\ 
            EVT        & 92.9  & 95.5  & 87.8  &  85.6         \\
            \bottomrule[0.5mm]
            \end{tabular}
    }
    \label{tab:threshold}
\vspace{-8pt}
\end{wraptable}
\textbf{Threshold selection.}
\label{app:threshold}
We investigate the effect of different thresholding strategies for anomaly score calibration. In the main experiments, we adopt an Extreme Value Theory (EVT) based strategy, where a Generalized Extreme Value (GEV) distribution is fitted to the image-level anomaly score set $\{s_i\}$ and the threshold is defined as:

\begin{equation}
    \tau = F_{\text{GEV}}^{-1}(0.98).
\end{equation}

Here, we additionally evaluate the commonly used k-sigma rule as an alternative thresholding strategy. 
The threshold $\tau$ is determined adaptively based on the mean \(\mu_s\) and standard deviation \(\sigma_s\) of the anomaly score distribution in the training set: $\tau = \mu_s + \kappa \cdot \sigma_s$ where $\kappa$ is a parameter that controls the strictness of the decision boundary, and is set to $\kappa=3$ in our experiments.
Unless otherwise specified, all experiments in the appendix use the k-sigma rule. As shown in Tab.~\ref{tab:threshold}, the k-sigma rule yields comparable performance to EVT, demonstrating that our method is robust to the choice of thresholding strategy.

\begin{table}[h]
\centering
\begin{minipage}{0.48\linewidth}
    \centering
    \caption{Ablation study on CAAS.}
    \resizebox{\linewidth}{!}{
        \begin{tabular}{lcccc}
        \toprule[0.5mm]
        \multirow{2}{*}{\textbf{Setting}} 
        & \multicolumn{2}{c}{\textbf{MVTec-AD}} 
        & \multicolumn{2}{c}{\textbf{VisA}} \\ 
        \cmidrule(lr){2-3} \cmidrule(lr){4-5} 
        & \textbf{Accuracy} & \textbf{F1} 
        & \textbf{Accuracy} & \textbf{F1} \\ 
        \midrule
        Baseline           & 93.1 & 95.6 & \textbf{88.3} & \textbf{85.7} \\ 
        Visual + Textual   & 94.2 & 96.4 & 85.8 & 82.5 \\
        Text               & 93.9 & 96.7 & 87.1 & 84.2 \\ 
        \rowcolor{gray!20} \textbf{Visual} 
                           & \textbf{94.6} & \textbf{96.7} & 86.7 & 83.6 \\
        \bottomrule[0.5mm]
        \end{tabular}
    }
    \label{tab:caas_ab}
\end{minipage}
\hfill
\begin{minipage}{0.49\linewidth}
    \centering
    \caption{Ablation study on layer range selection.}
    \resizebox{\linewidth}{!}{
        \begin{tabular}{lcccc}
        \toprule[0.5mm]
        \multirow{2}{*}{\textbf{Layer Range}} 
        & \multicolumn{2}{c}{\textbf{MVTec-AD}} 
        & \multicolumn{2}{c}{\textbf{VisA}} \\
        \cmidrule(lr){2-3} \cmidrule(lr){4-5}
        & \textbf{Accuracy} & \textbf{F1} 
        & \textbf{Accuracy} & \textbf{F1} \\
        \midrule
        1--8   & 94.0 & 96.1 & 86.7 & 83.4\\
        9--14  & \textbf{94.6} & \textbf{96.7} & 86.7 & 83.6\\
        15--21 & 91.6 & 95.9 & 80.6 & 75.7\\
        22--28 & 93.8 & 93.1 & 86.3 & 83.1\\
        \bottomrule[0.5mm]
        \end{tabular}
    }
    \label{tab:layer}
\end{minipage}

\end{table}

\subsection{Further Analyses and Results on  Confidence-Aware Attention Sharpening }
\label{appendix_caas}

\textbf{Impact of CAAS.}
Tab.~\ref{tab:caas_ab} evaluates the effect of CAAS by amplifying attention to visual tokens ($\alpha=0.6$). The results indicate that strengthening visual attention contributes to improved performance. We further examine suppressing attention to erroneous textual priors ($\beta=-0.4$), as well as jointly modulating both visual and textual attention. While all these strategies outperform the baseline, visual-only amplification yields the most effective improvement.  



\begin{wrapfigure}{r}{0.4\textwidth} 
  \centering
   \vspace{-10pt}
  \includegraphics[width=\linewidth]{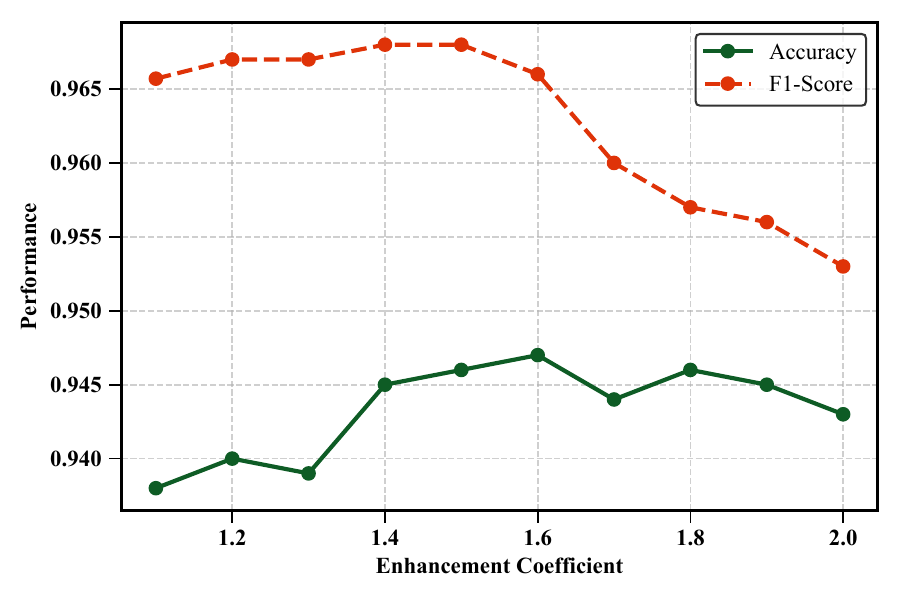}
  \vspace{-10pt}
  \caption{Ablation study of scaling factor $a$ on the MVTec. }
  \label{acc_f}
\vspace{-25pt}
\end{wrapfigure}
\begin{figure*}[h]
\centering

\begin{minipage}{0.78\linewidth}
    \centering
    \includegraphics[width=\linewidth]{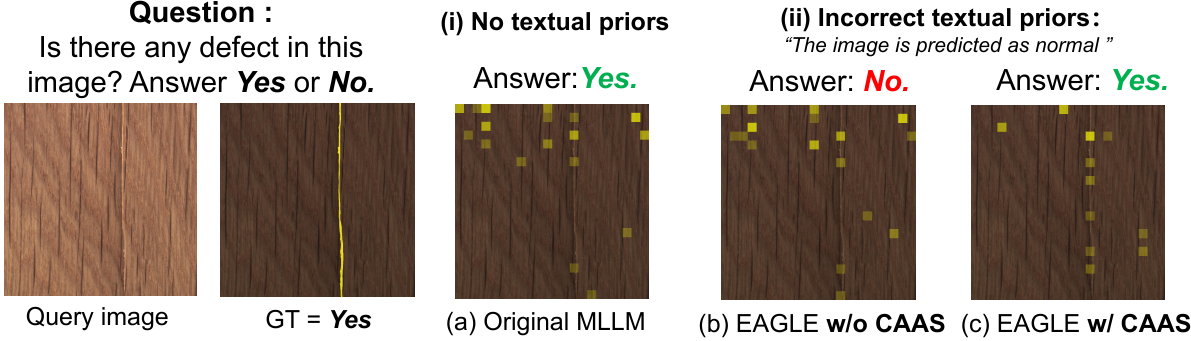}
    
    \captionof{figure}{
    Attention map visualizations corresponding to the layer-wise prediction dynamics.
    }
    \label{fig:11}
\end{minipage}

\vspace{10pt}

\begin{minipage}{0.95\linewidth}
\centering

\begin{subfigure}{0.32\linewidth}
    \centering
    \includegraphics[width=\linewidth]{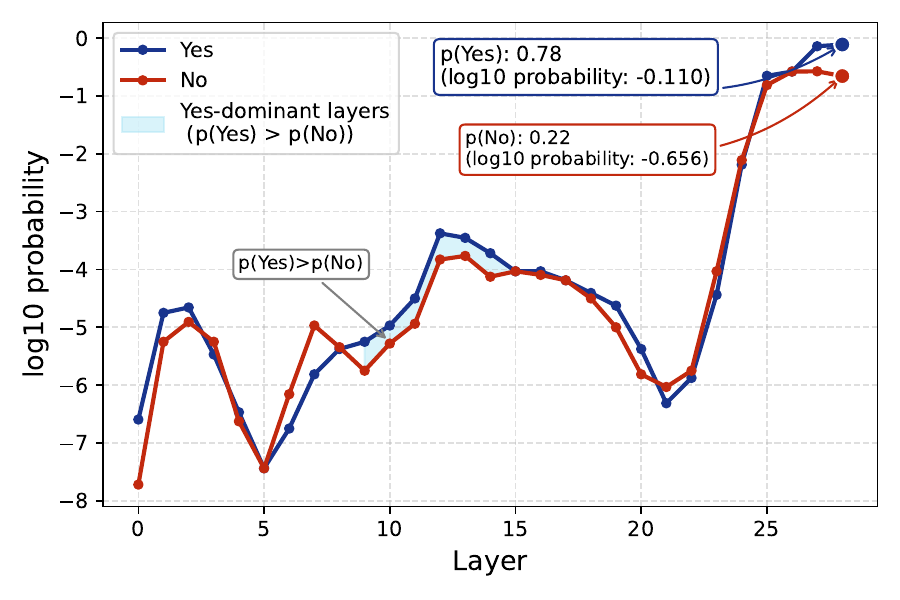}
    \caption{Original MLLM}
\end{subfigure}
\hfill
\begin{subfigure}{0.32\linewidth}
    \centering
    \includegraphics[width=\linewidth]{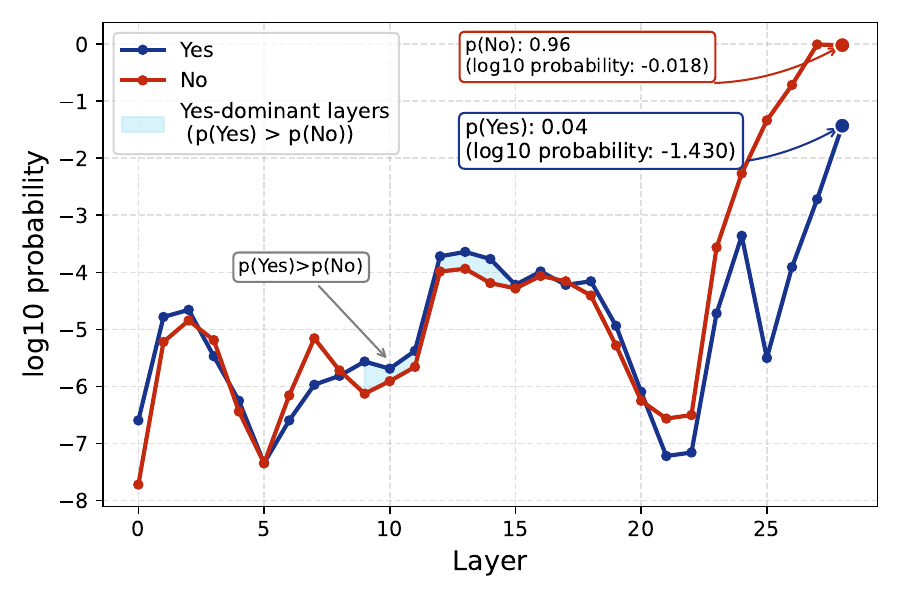}
    \caption{EAGLE w/o CAAS}
\end{subfigure}
\hfill
\begin{subfigure}{0.32\linewidth}
    \centering
    \includegraphics[width=\linewidth]{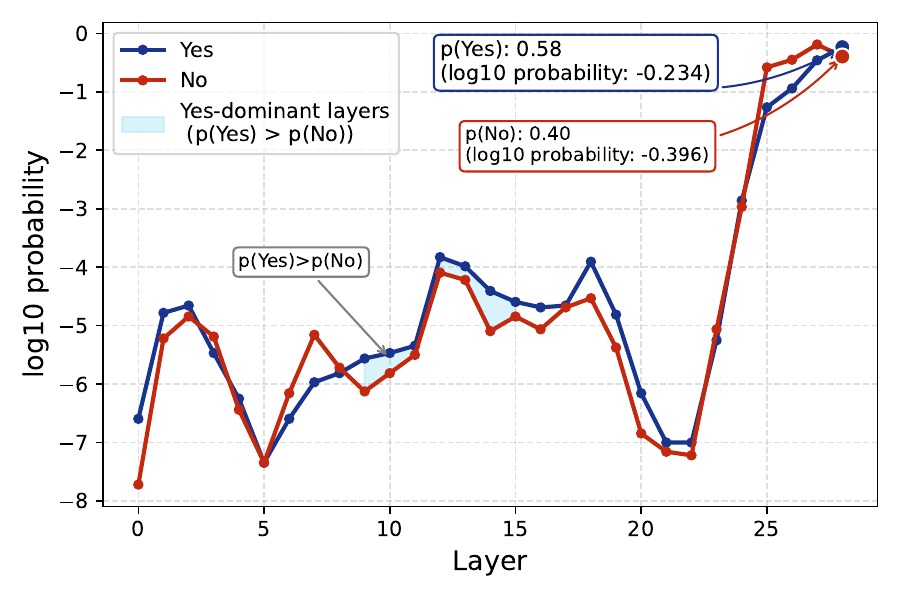}
    \caption{EAGLE w/ CAAS}
\end{subfigure}

\captionof{figure}{
Layer-wise log probability evolution of candidate answers (``\textit{Yes}'' and ``\textit{No}'') under incorrect textual priors.
}
\label{fig:12}

\end{minipage}
\end{figure*}

\textbf{The influence of the target layer.}
In this work, the selection of the layer range for CAAS (layers 9–14) is supported by both prior studies and empirical analysis. Existing works\cite{yin2025clearsight, chen2025spatial, zhang2025mllms} consistently indicate that intermediate layers in MLLMs play a crucial role in visual processing and cross-modal fusion. Based on this, we further conduct ablation studies across different layer ranges, as shown in Tab.~\ref{tab:layer}. The results show that applying attention modulation in intermediate layers (9–14) yields the most consistent performance improvement, whereas applying it to earlier or deeper layers leads to limited gains, and even causes performance degradation in later layers (15–22). We further illustrate the effect of intermediate layers on correct prediction through a case study in Fig.~\ref{fig:12}.

\label{app:alpha}

Furthermore, we conduct ablation studies on scaling factor $\alpha$ to investigate its impact on model performance. As shown in Figure~\ref{acc_f}, increasing $\alpha$ within a moderate range ($0 < \alpha < 1$) leads to continuous performance improvement, demonstrating that moderately enhancing visual attention can effectively mitigate hallucinations caused by erroneous linguistic priors. However, when $\alpha$ exceeds 0.6, performance begins to decline. This trend is likely caused by excessive amplification of visual features disrupting the balance between visual and linguistic information. These observations further support our core design principle that visual and linguistic modalities should be dynamically balanced rather than excessively biased toward a single modality. Moderate visual enhancement enables the model to correct unreliable linguistic priors, while excessive intervention disrupts the original multimodal reasoning structure.

\label{app:layer}






\textbf{Layer-wise Prediction Dynamics under Incorrect Textual Priors.}
Fig.~\ref{fig:12} illustrates the layer-wise evolution of log probabilities for 
candidate answers (\textit{Yes} and \textit{No}) under incorrect textual priors across 
different settings, with results obtained on the wood category of MVTec-AD. 
Fig.~\ref{fig:11}presents the corresponding attention map visualizations, where the 
attention maps are extracted from the 27th Transformer layer of the MLLM.

In the original MLLM (Fig.~\ref{fig:12}(a)), the model ultimately produces the 
correct prediction with a relatively high probability ($p(\text{Yes})=0.78$), indicating 
that without any prompt intervention, the MLLM is capable of identifying anomalies based 
solely on its own visual understanding.

In EAGLE without CAAS (Fig.~\ref{fig:12}(b)), the erroneous textual prior provided 
by the expert model (\textit{"The image is predicted as normal"}) misleads the model, 
resulting in an incorrect prediction with an extremely low probability ($p(\text{Yes})=0.04$, 
Answer: \textit{No}). The corresponding attention map reveals a marked decrease in the 
model's focus on the true defect region, confirming that the incorrect textual prior 
suppresses visual evidence.

In EAGLE with CAAS (Fig.~\ref{fig:12}(c)), CAAS modulates the attention weights to 
amplify the model's focus on visual tokens, gradually recovering the probability of 
\textit{Yes} to $p(\text{Yes})=0.58$ and yielding the correct prediction (Answer: 
\textit{Yes}). The attention map shows that the model refocuses on the true defect region, 
validating the robustness and effectiveness of CAAS against erroneous textual priors.

\begin{figure*}[h]    
  \centering    
  \subfloat[Correct vs incorrect attention ]   
  {
  \label{fig:subfig1}\includegraphics[width=0.49\textwidth]{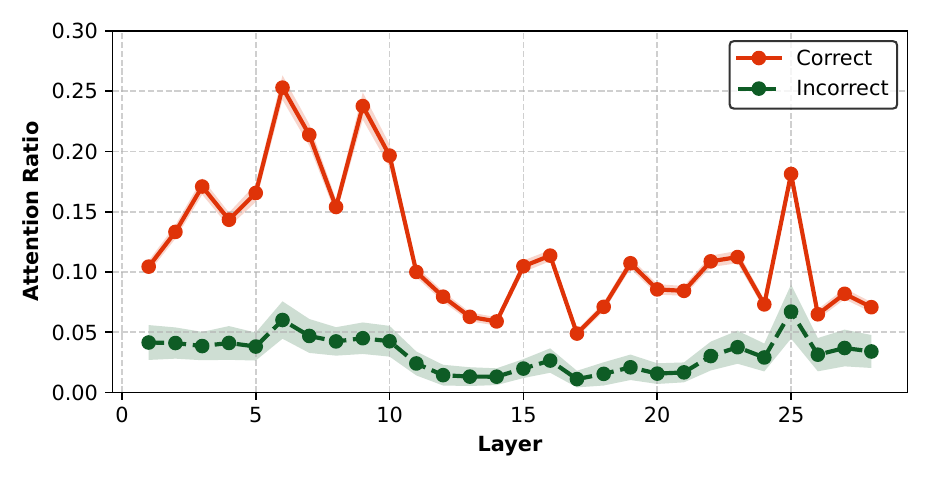}
  }
  \subfloat[EAGLE vs baseline attention.]
  {
    \label{fig:subfig2}\includegraphics[width=0.49\textwidth]{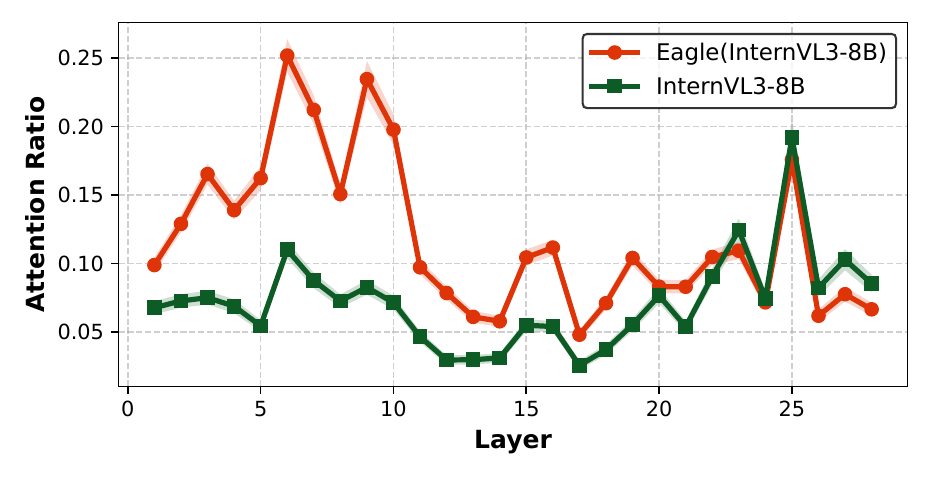}
  }
  \caption{Attention ratios to ground-truth anomalous regions across Transformer layers.}   
  \label{fig:app-attention}      
  
\end{figure*}
\subsection{Further Analysis}
\label{app:attention}
Fig.~\ref{fig:app-attention} presents the attention ratio analysis for InternVL3-8B, 
consistent with the findings observed on Qwen2.5-VL-7B. In Fig.~\ref{fig:app-attention}~(a), correctly predicted samples exhibit substantially higher attention ratios than 
incorrect ones across most layers, further confirming the strong correlation 
between detection accuracy and attention to anomalous regions. In 
Fig.~\ref{fig:app-attention}~(b), EAGLE consistently assigns higher attention ratios 
compared to the standalone InternVL3-8B, demonstrating that the effectiveness 
of dual-modality prompts in guiding visual attention generalizes across 
different MLLM architectures.

\section{Limitations}
\label{sec:limtations}
\textbf{Sensitivity to Distribution Shift.} The TGPS mechanism relies on normal samples from 
the training set for threshold estimation, which requires the dataset to conform to an extreme 
value distribution. When test samples exhibit significant distribution shift, i.e., when they 
differ substantially from the training samples in terms of object quantity, viewpoint, color, 
or object morphology, the estimated threshold may become unreliable, generating erroneous 
textual priors that exacerbate MLLMs hallucinations rather than mitigate them. Future work 
should explore adaptive thresholding strategies to handle distribution shifts. A more detailed 
discussion is provided in Appendix~\ref{app:AS}.

\textbf{Dependency on MLLMs.} The performance of EAGLE is inherently bounded by the 
capabilities of the underlying MLLMs backbone. Issues such as inadequate visual grounding, 
limited reasoning capacity, or inherent biases in the pretrained model directly affect the 
effectiveness of our framework. In particular, when the MLLMs perform poorly on a given 
dataset, the detection capability of EAGLE may become heavily dependent on the performance 
of the expert model. A detailed analysis and discussion of this phenomenon is provided in 
Appendix~\ref{app:addition_results}. As MLLMs architectures continue to evolve, we believe the performance of EAGLE will improve correspondingly.

\textbf{Limitations of Visual Prompt Annotation.} In the current implementation, visual 
prompts are provided by annotating anomalous regions with red bounding boxes derived from 
segmentation results. However, this annotation approach has inherent precision limitations, 
and more accurate annotation methods are expected to further enhance the guidance effect of 
visual prompts. A more detailed discussion is provided in Appendix~\ref{app:prompt}.

\section{Prompt detail}
\subsection{Prompt Design Details}
\label{app:prompt}

All experiments in this paper follow a unified prompting scheme to ensure fair and consistent evaluation across different MLLM backbones. The prompt is constructed by combining a fixed instruction, optional expert-generated cues, and a question requiring a binary decision.

\textbf{System Instruction.}
You are my industrial image inspection assistant. You will receive multiple images simultaneously, 
including a template image, a query image, and a query image with red bounding boxes. Based on the input images and the accompanying textual information, answer the given question. The question is multiple-choice. Respond only with the letter of the correct option (e.g., A, B, C, or D). Do not include explanations or extra text.

\begin{figure}[htbp]
  \vskip 0.2in
  \begin{center}
    \centerline{\includegraphics[width=0.9\columnwidth]{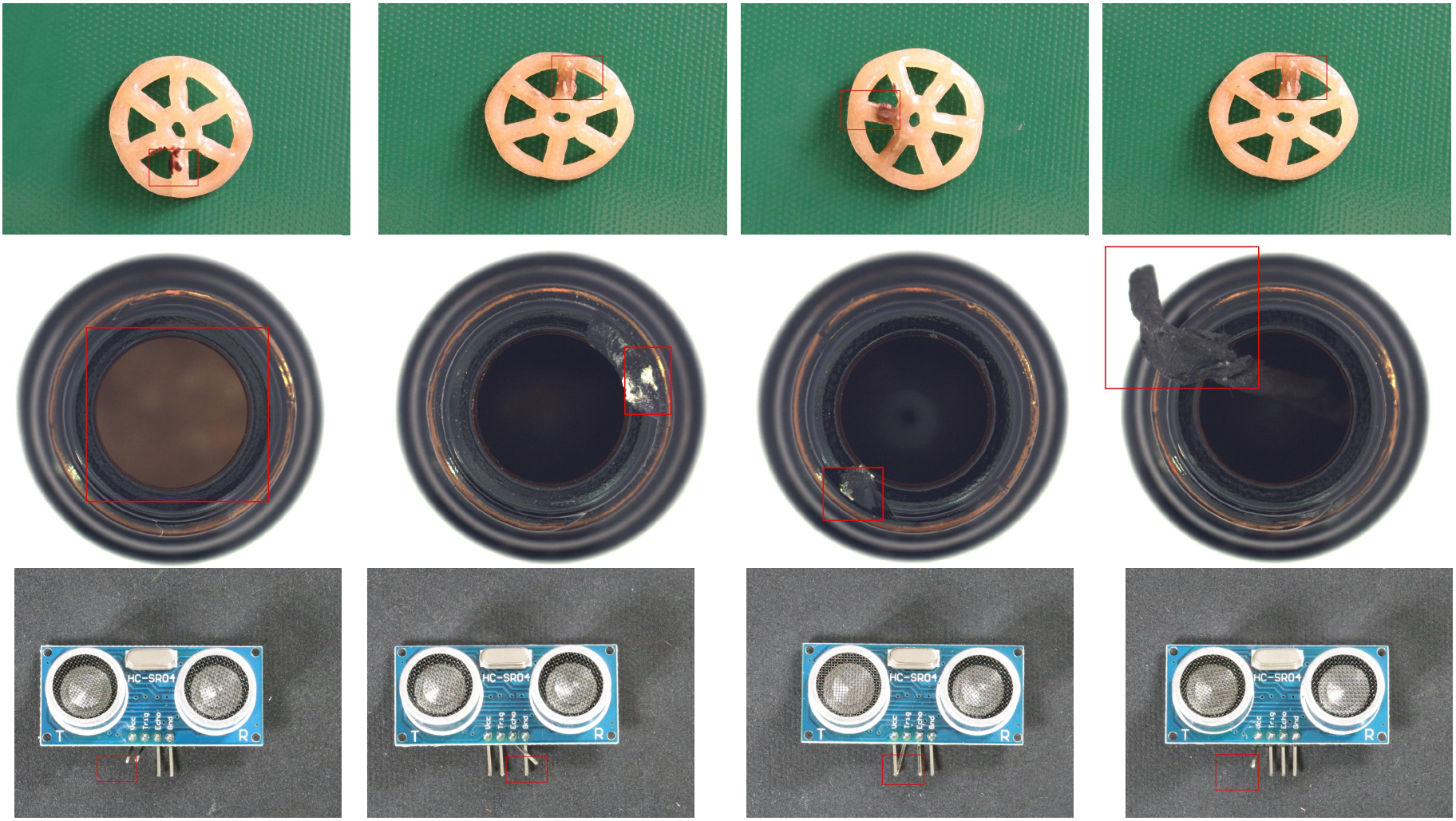}}
    \caption{Visual prompt}
    \label{fig:visualprompt}
  \end{center} 
\end{figure}

\textbf{Expert-Guided Prompts.}
When available, expert model outputs are incorporated into the prompt in two forms:
\begin{itemize}
\item \textbf{Textual Prompt.}
  \begin{itemize}
    \item \textbf{Abnormal Prompt.}
    The query image is predicted as anomalous. The position of the red bounding box on the query image is the predicted defect location. Answer with the option's letter from the given choices directly! Is there any defect in the object? A. Yes. B. No.
    \item \textbf{Normal Prompt.}
    The query image is predicted as normal. Answer with the option's letter from the given choices directly! Is there any defect in the object? A. Yes. B. No.
  \end{itemize}
\item \textbf{Visual Prompts.} see Fig.~\ref{fig:visualprompt}
\end{itemize}
To avoid misleading cues, visual prompts are provided only for images classified as anomalous by the expert model.

\label{app:sampled}

\begin{wraptable}{r}{0.5\textwidth}
\centering
\vspace{-10pt}
\caption{Localization quality of visual prompts.}
\label{tab:loc_quality}
\begin{tabular}{lcc}
\toprule[0.5mm]
\textbf{Dataset} & \textbf{Mean IoU} & \textbf{IoU@0.5} \\
\midrule
MVTec-AD & 0.38 & 3.9 \\
VisA     & 0.16 & 5.5 \\
\bottomrule[0.5mm]
\end{tabular}
\end{wraptable}

\subsection{Visual Prompt Generation}
In this work, visual prompts are generated by the expert model based on the anomaly map. Specifically, we first upsample the patch-level anomaly scores to the original image resolution, resulting in $\tilde{A} \in \mathbb{R}^{H \times W}$, and normalize it as:
\[
\hat{A}(x,y)=\frac{\tilde{A}(x,y)}{\max_{u,v}\tilde{A}(u,v)+\epsilon},
\]
where $\epsilon$ is a small constant to ensure numerical stability. This normalization maps the anomaly responses to a unified scale for subsequent processing.

We then apply adaptive thresholding to binarize the anomaly map and extract candidate anomalous regions. For each connected component, we compute its bounding box $b_i=(x_i^{(1)},y_i^{(1)},x_i^{(2)},y_i^{(2)})$, and assign a confidence score defined as the maximum anomaly response within the region.

To account for fragmented or overlapping responses, we adopt a lightweight merging strategy. For any two bounding boxes $b_i$ and $b_j$, their Intersection-over-Union (IoU) is defined as
\[
\mathrm{IoU}(b_i, b_j)=\frac{|b_i \cap b_j|}{|b_i \cup b_j|},
\]
and the center of a bounding box is given by
\[
c(b_i)=\left(\frac{x_i^{(1)}+x_i^{(2)}}{2},\,\frac{y_i^{(1)}+y_i^{(2)}}{2}\right),
\]
with the corresponding center distance defined as
\[
d(b_i,b_j)=\|c(b_i)-c(b_j)\|_2.
\]
We group bounding boxes that satisfy
\[
\mathrm{IoU}(b_i, b_j) > \tau_{\mathrm{iou}} \quad \text{or} \quad d(b_i, b_j) < \tau_{\mathrm{dist}},
\]
and merge them by taking the union of their spatial extents. Here, $\tau_{\mathrm{iou}}$ and $\tau_{\mathrm{dist}}$ denote the IoU and distance thresholds, respectively. The merged boxes are then ranked by their anomaly scores, and the top-$K$ boxes are selected as visual prompts (we use $K=3$ in this work).

Tab.~\ref{tab:loc_quality} reports the localization quality of the generated visual prompts. As shown, the Mean IoU is relatively low (e.g., 0.38 on MVTec-AD), and the proportion of boxes with IoU@0.5 is also limited. This indicates that the proposed method does not aim for precise localization, but rather provides coarse spatial coverage of anomalous regions.

This also reflects a limitation of our approach. Future work may explore more accurate anomaly localization and bounding box generation strategies, such as incorporating pixel-level segmentation or learning-based post-processing methods, to further improve localization quality while maintaining efficiency.

\end{document}